\documentclass[lettersize,journal]{IEEEtran}
\pdfoutput=1

\usepackage[table]{xcolor}
\usepackage{amsmath,amsfonts}
\usepackage{algorithmic}
\usepackage{algorithm}
\usepackage{array}
\usepackage[caption=false,font=normalsize,labelfont=sf,textfont=sf]{subfig}
\usepackage{textcomp}
\usepackage{stfloats}
\usepackage{url}
\usepackage{verbatim}
\usepackage{graphicx}
\usepackage[numbers,sort&compress]{natbib}

\usepackage{hyperref}
\usepackage{color}

\usepackage{diagbox}
\usepackage{makecell}

\usepackage{tikz}
\usetikzlibrary{mindmap}
\usepackage{smartdiagram}
\usesmartdiagramlibrary{additions}
\usepackage{forest}
\usetikzlibrary{shadows}
\usetikzlibrary{patterns}
\usepackage{amsmath,amssymb,amsfonts} 
\usepackage{algorithmic} 
\usepackage{textcomp} 
\usepackage{color}

\usepackage{hyperref}
\usepackage{cleveref}
\crefname{figure}{Fig.}{Fig.}
\usetikzlibrary{snakes}
\usepackage{chronosys}

\usepackage{nicematrix,booktabs}       

\usepackage[normalem]{ulem} %
\usepackage[switch]{lineno}

\newif\ifcleanversion
\cleanversiontrue  %
\usepackage{xspace}
\usepackage{soul} %
\soulregister\cite7 %
\soulregister\ref1 %
\soulregister\eqref1 %
\soulregister\~1 %

\ifcleanversion
    \newcommand{\redsout}[1]{} %
    \newcommand{\rev}[1]{\textcolor{black}{#1}}
    \newcommand{\revmath}[1]{\mathcolor{black}{#1}}
    \newcommand{\revsec}[1]{%
      \begingroup%
        \color{black}%
        #1%
      \endgroup%
    }
\else

    \newcommand\redsout[1]{%
      \begingroup%
      \setstcolor{red}%
      \setul{0.5ex}{1.2pt}
      \st{#1}%
      \endgroup%
    }%
    \newcommand{\rev}[1]{\textcolor{blue}{#1}}
    \newcommand{\revmath}[1]{\mathcolor{blue}{#1}}
    \newcommand{\revsec}[1]{%
      \begingroup%
        \color{blue}%
        #1%
      \endgroup%
    }
    
\fi

\usepackage{bm}
\usepackage{enumitem}

\usepackage{calligra}

\definecolor{lightcoral}{rgb}{0.94, 0.5, 0.5}
\definecolor{lightgreen}{rgb}{0.56, 0.93, 0.56}
\definecolor{lightblue}{rgb}{0.56, 0.56, 0.93}
\definecolor{lightyellow}{rgb}{0.94, 0.84, 0.6}
\definecolor{harvestgold}{rgb}{0.85, 0.57, 0.0}
\definecolor{brightlavender}{rgb}{0.75, 0.58, 0.89}
\definecolor{capri}{rgb}{0.0, 0.75, 1.0}
\definecolor{carminepink}{rgb}{0.92, 0.3, 0.26}
\definecolor{celadon}{rgb}{0.67, 0.88, 0.69}
\definecolor{darkpastelgreen}{rgb}{0.01, 0.75, 0.24}

\definecolor{lightblue}{rgb}{0.9, 0.9, 1}

\usepackage{booktabs, cellspace, tabularx}

\usepackage{nicematrix}

\newtheorem{definition}{Definition}
\newtheorem{challenge}{Challenge}
\newtheorem{scenario}{Scenario}

\newtheorem{remark}{Remark}

\newcommand{\challengehyperref}[1]{%
\hyperref[#1]{C\getrefnumber{#1}}%
}
\newcommand{\challengeName}[1]{%
  \ifcase#1\relax%
    multimodal catastrophic forgetting%
  \or%
    modality imbalance%
  \or%
    complex modality interaction%
  \or%
    high computational costs%
  \or%
    degradation of pre-trained zero-shot capability%
  \else%
    \PackageError{challengeName}{Number out of range: #1}{}%
  \fi%
}

\newcommand{\challengeNumRef}[2][]{%
  \ifcase#2\relax%
    \hyperlink{item: mmcl forgetting}{0}#1%
  \or%
    \hyperref[item:Modality imbalance]{1}#1%
  \or%
    \hyperref[item:Modality interaction]{2}#1%
  \or%
    \hyperref[item:High computational costs]{3}#1%
  \or%
    \hyperref[item:Forgetting of the pre-trained knowledge]{4}#1%
  \else%
    \PackageError{challengeNumRef}{Number out of range: #2}{}%
  \fi%
}

\NewDocumentCommand{\challengeNumDesc}{s m}{%
  \IfBooleanTF{#1}
    {\challengeNumRef{#2}: \challengeName{#2}}%
    {Challenge~\challengeNumRef{#2}: \challengeName{#2}}%
}

\NewDocumentCommand{\challengeNum}{s m}{%
  \IfBooleanTF{#1}
    {\challengeNumRef{#2}}%
    {Challenge~\challengeNumRef{#2}}%
}

\NewDocumentCommand{\challengeNumTwo}{s m m}{%
  \IfBooleanTF{#1}
    {\challengeNumRef{#2} and \challengeNumRef{#3}}%
    {Challenges~\challengeNumRef{#2} and \challengeNumRef{#3}}%
}

\usepackage{multirow}
\usepackage{pifont}%
\newcommand{\cmark}{\ding{51}}%
\newcommand{\starMark}{\ding{76}}%
\newcommand{\challengeMark}{$\bullet$}%

\crefname{section}{\S}{\S\S}

\newcommand{\climb}{CLiMB}
\newcommand{\taskSeq}{TS}

\newcommand{\mmclMethods}{\textbf{MMCL Methods. }}
\newcommand{\summary}{\textbf{Summary with Pros \& Cons. }}

\def\Put(#1,#2)#3{\leavevmode\makebox(0,0){\put(#1,#2){#3}}}

\makeatletter
    \newcommand{\linebreakand}{%
      \end{@IEEEauthorhalign}
      \hfill\mbox{}\par
      \mbox{}\hfill\begin{@IEEEauthorhalign}
    }
    \makeatother

\newcommand{\usePattern}{0}  %

\newcommand{\smartPatternParams}[1]{
    text width=#1,
    text=black, 
    \ifnum\usePattern=1
        pattern=checkerboard,
        pattern color=gray,
    \fi
}

\newcommand{\nodeParamsPattern}[3]{
    text width={#1},
    color=#2,
    fill=#3,
    text=black,
    \ifnum\usePattern=1
        pattern=checkerboard,
        pattern color=#3,
    \fi
}
\newcommand{\nodeParamsPatternWidthSndLevel}{2.2cm}

\begin{document}
\bstctlcite{IEEEexample:BSTcontrol}

\ifcleanversion
\else
    \linenumbers   
\fi

\title{Recent Advances of Multimodal Continual Learning: A Comprehensive Survey}

\author{Dianzhi Yu, Xinni Zhang, Yankai Chen, Aiwei Liu, Yifei Zhang, Philip S. Yu,~\IEEEmembership{Fellow,~IEEE}, \\and Irwin King,~\IEEEmembership{Fellow,~IEEE}
\thanks{Dianzhi Yu, Xinni Zhang, Yankai Chen, Yifei Zhang, and Irwin King are with the Department of Computer Science and Engineering, The Chinese University of Hong Kong, Hong Kong, China (E-mail: \{dzyu23, xnzhang23, ykchen, yfzhang, king\}@cse.cuhk.edu.hk). 
Aiwei Liu is with the School of Software, Tsinghua University, Beijing 100084, China (E-mail: liuaw20@mails.tsinghua.edu.cn).
Philip S. Yu is with the Department of Computer Science, University of Illinois Chicago, Chicago, IL 60607 USA (e-mail: psyu@uic.edu).}%
}

\markboth{Journal of \LaTeX\ Class Files,~Vol.~14, No.~8, August~2021}%
{Shell \MakeLowercase{\textit{et al.}}: A Sample Article Using IEEEtran.cls for IEEE Journals}

\IEEEpubid{0000--0000/00\$00.00~\copyright~2021 IEEE}

\IEEEtitleabstractindextext{%
\begin{abstract}
Continual learning (CL) aims to empower machine learning models to learn continually from new data, while building upon previously acquired knowledge without forgetting.
As models have evolved from small to large pre-trained architectures, and from supporting unimodal to multimodal data, multimodal continual learning (MMCL) methods have recently emerged. 
\rev{The primary complexity of MMCL is that it extends beyond a simple stacking of unimodal CL methods.
Such straightforward approaches often suffer from multimodal catastrophic forgetting, yielding unsatisfactory performance. 
In addition, MMCL introduces new challenges that unimodal CL methods fail to adequately address, including modality imbalance, complex modality interaction, high computational costs, and degradation of pre-trained zero-shot capability of multimodal backbones.}
In this work, we present the \textit{first} comprehensive survey on MMCL. We provide essential background knowledge and MMCL settings, as well as a structured taxonomy of MMCL methods.
We categorize MMCL methods into four categories, i.e., regularization-based, architecture-based, replay-based, and prompt-based methods, explaining their methodologies and highlighting their key innovations. 
Additionally, to prompt further research in this field, we summarize open MMCL datasets and benchmarks, 
\rev{provide an in-depth discussion},
and discuss several promising future directions. 
We have also created a GitHub repository for indexing relevant MMCL papers and open resources available at \url{https://github.com/LucyDYu/Awesome-Multimodal-Continual-Learning}.

\end{abstract}

}

\maketitle

\IEEEdisplaynontitleabstractindextext
\IEEEpeerreviewmaketitle

\section{Introduction}
\label{sec: Introduction}

In recent years, machine learning (ML) has achieved significant advancements, contributing to the resolution of a wide range of practical problems. 
In conventional settings, most ML models operate within the so-called ``single-episode" paradigm, being trained on \textit{static} and \textit{single} datasets, while evaluated under the independent and identically distributed (i.i.d.) assumption~\cite{Hadsell2020Embracing}.
However, this ``single-episode" paradigm may not equip the trained models with the capability to adapt to new data or perform new tasks, failing to align with the aspiration of developing intelligent agents for dynamically evolving environments. 
To address this issue, the ML community is motivated to develop \textit{continual learning (CL)}, also known as lifelong learning or incremental learning, which trains models incrementally on new tasks and maintains early knowledge without requiring full-data retraining~\cite{chen2022lifelong, kudithipudi2022biological, vandeVen2022Three, Wang2024Comprehensive}.
CL has become increasingly important in various fields and applications, including but not limited to the Internet-of-Things~\cite{LealFilho2024Continual}, website detection~\cite{Ejaz2023Lifelong}, and activity recognition~\cite{Hu2018Novel}.

\IEEEpubidadjcol
The main challenge of CL is \textit{catastrophic forgetting}: a phenomenon that when tasks are trained sequentially, training on the new task greatly disrupts performance on previously learned tasks~\cite{McCloskey1989Catastrophic,Ratcliff1990Connectionist}, as unconstrained fine-tuning drives parameters moving far from the old optimal state~\cite{Hassabis2017NeuroscienceInspired}. 
CL aims to develop learning systems that achieve a balance between \textit{plasticity} for continuous knowledge acquisition and \textit{stability} for retaining previously learned information~\cite{Mermillod2013Stabilityplasticity, Masana2022ClassIncremental}.
\rev{Such a process essentially imitates the cognitive flexibility observed in brains, which continually learn diverse skills throughout the human lifespan~\cite{qu2023recentadvancescontinuallearning}.}
By enabling models to adapt to new tasks without forgetting, CL offers advantages in terms of resource and time efficiency compared to the traditional approach of exhaustive model retraining on full task datasets. 
Furthermore, due to issues of storage limitations, privacy concerns, etc., the potential inaccessibility of historical training data makes full-data training unfeasible, highlighting the efficiency and effectiveness of CL in memorizing former knowledge and acquiring up-to-date one from dynamic environments.

\rev{Despite significant progress in CL, most efforts have been devoted to a single data modality, such as vision~\cite{Li2017Learning, Kirkpatrick2017Overcoming, Rebuffi_2017_CVPR, mallya2018packnet, Gao2022Efficient, Liu2023Model}, language~\cite{Wang2023Orthogonal, Wang2024InsCL, Wu2024FMALLOC, Ho2024PrototypeGuided, Yang2024MoRAL}, or audio~\cite{Wang2019Continual, Ma2021Continual}.}
This \textit{unimodal focus} overlooks the multimodal nature of real-world environments, which are inherently complex and composed of diverse modalities rather than a single one.
With the rapid growth of such multimodal data, e.g., data proliferation of images, texts, and videos on platforms like Meta and TikTok, it is imperative to develop AI systems capable of learning continually from \textit{multimodal sources}, hence the rise of the \textbf{multimodal continual learning (MMCL)} setting.
These MMCL systems need to effectively integrate and process various multimodal data streams~\cite{baltrusaitis2017multimodalmachinelearningsurvey, Wu2023Multimodal} while managing to preserve previously acquired knowledge. 
More importantly, this MMCL setting better mimics the process of learning and integrating information across different modalities in human biological systems, ultimately enhancing the overall perception and cognitive capabilities when dealing with real-world complexities~\cite{mroczko2016perception, Sarfraz2024Unimodal}. 
Illustrations of unimodal CL and MMCL are provided in \Cref{fig: non-tech Illustration of CL and MMCL}.

\newcommand{\mySizeNonTech}{0.48\linewidth}

\begin{figure*}[!t]
\centering
\hspace{-0.35cm} %
\subfloat[\small{Unimodal CL}]{
    \label{fig: non-tech illustration Unimodal CL}
    \includegraphics[width=\mySizeNonTech,page=19]{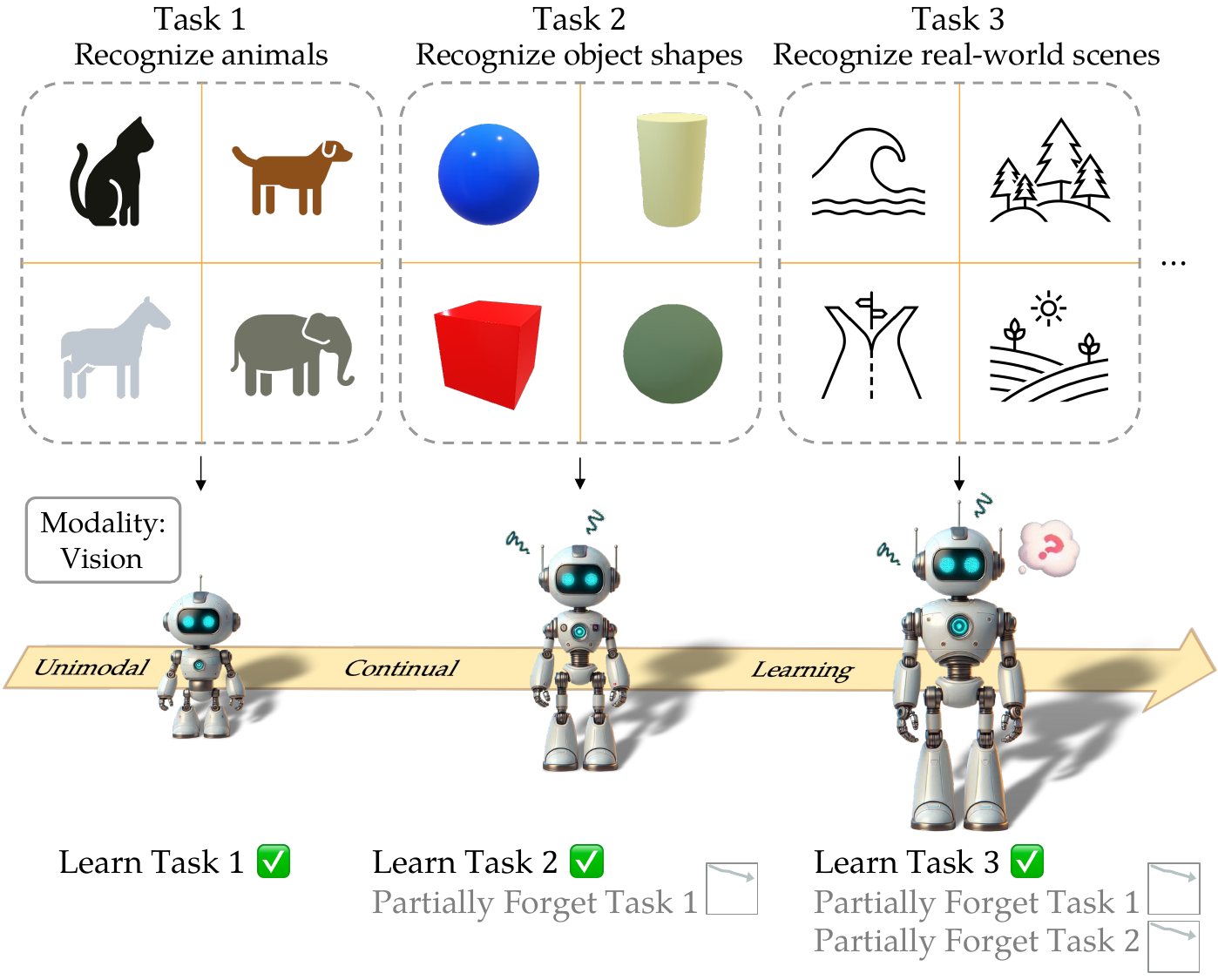}
}
\hspace{-0.15cm} %
\subfloat[\small{\rev{Multimodal CL}}]{
    \label{fig: non-tech illustration MMCL}    
    \includegraphics[width=\mySizeNonTech,page=20]{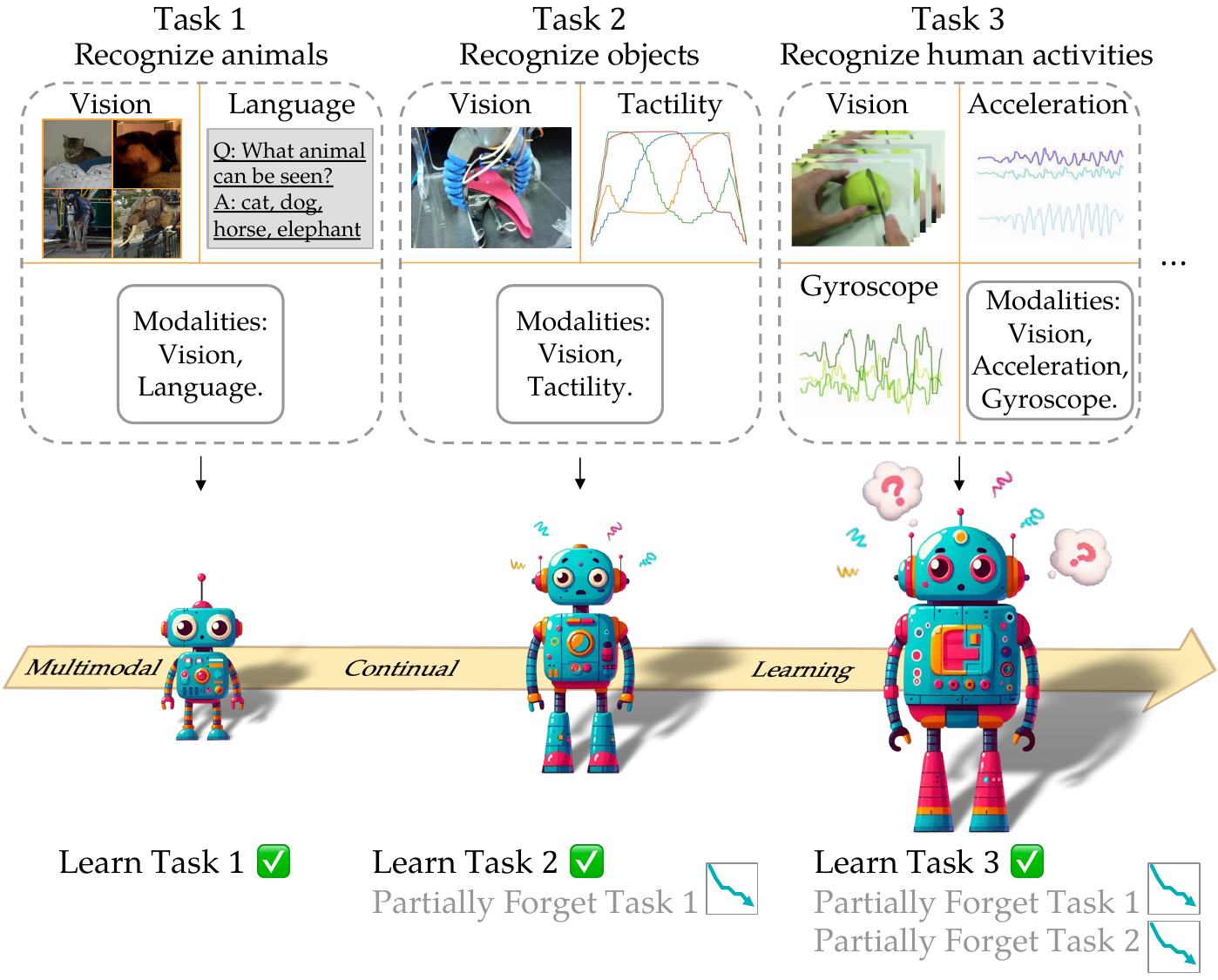}  
}
\caption{Graphical illustrations of CL and MMCL. (a) Unimodal CL. The model continually learns new tasks. While learning a new task, the model tends to forget the previously learned tasks. CL aims to mitigate forgetting. (b) Multimodal CL. In the multimodal setting, the model continually learns new tasks, and the dataset is multimodal. Forgetting in MMCL tends to be more severe due to challenges mentioned in Section~\texorpdfstring{\protect\hyperlink{mylink challenges}{\getrefnumber{sec: Introduction}}{}}~.
Example tasks in \Cref{fig: non-tech illustration Unimodal CL} are adapted based on SCD \cite{Lao2023MultiDomain}, VQACL \cite{Zhang2023VQACL}, CLEVR \cite{Johnson2017CLEVR} and GQA \cite{Hudson2019GQA}.
\rev{Example tasks in \Cref{fig: non-tech illustration MMCL} are adapted based on SCD \cite{Lao2023MultiDomain}, VQACL \cite{Zhang2023VQACL}, exFeCAM~\cite{Sun2021Multimodal} and CMR-MFN~\cite{Wang2023Confusion}.}
}
\label{fig: non-tech Illustration of CL and MMCL}
\end{figure*}

\renewcommand{\aboverulesep}{0pt}
\newcommand{\mySizeChallenges}{-0.3cm}
\begin{figure}[!t]

    \subfloat[\small{\rev{Model components and challenges}}]{
        \begingroup  %
        \hspace{\mySizeChallenges} %
        \centering
        \includegraphics[height=5.2cm,page=17]{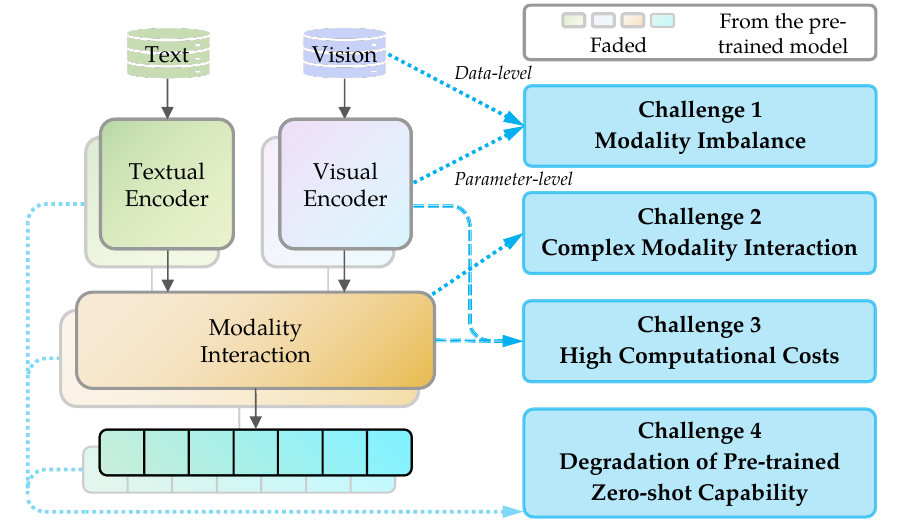}
        \Put(-49,236){\hyperref[item:Modality imbalance]{ \ }{\tikz \fill [opacity=0] (0,0) rectangle (0.5,0.25) ;}}
        \Put(-52,176){\hyperref[item:Modality interaction]{ \ }{\tikz \fill [opacity=0] (0,0) rectangle (0.5,0.25) ;}}
        \Put(-56,116){\hyperref[item:High computational costs]{ \ }{\tikz \fill [opacity=0] (0,0) rectangle (0.5,0.25) ;}}
        \Put(-59,56){\hyperref[item:Forgetting of the pre-trained knowledge]{ \ }{\tikz \fill [opacity=0] (0,0) rectangle (0.5,0.25) ;}}
        \endgroup  %
    
    }
    
    \subfloat[\small{\rev{Summary of challenge-component relationships}}]{
        \hspace{0.3cm} %
        \centering
        \begingroup  %
        \renewcommand{\arraystretch}{0.9}  %
        \hspace{\mySizeChallenges} %
        \hspace{\mySizeChallenges} %
        \hspace{\mySizeChallenges} %
        \setlength{\tabcolsep}{2.1pt}
        \begin{NiceTabular}{w{c}{2.2cm}|c|c|c|c}
            \CodeBefore
            \rowcolors{1}{white}{lightblue}
            \Body
        
            \toprule
            \diagbox{{Challenges\ \ }}{\raisebox{-2.4ex}{Comp.\ }}   & \makecell{ Multimodal\\ Input Data}&  \makecell{Modality \\ Encoders} & \makecell{Modality\\ Interaction} & Output\\
            \midrule
            \challengehyperref{item:Modality imbalance} & \challengeMark &  \challengeMark &  & \\
            \challengehyperref{item:Modality interaction}  & & & \challengeMark & \\
            \challengehyperref{item:High computational costs}  & & \challengeMark & \challengeMark & \\
            \challengehyperref{item:Forgetting of the pre-trained knowledge} & &  \challengeMark & \challengeMark & \challengeMark\\
            \bottomrule
          \end{NiceTabular}
          
        \endgroup  %
    }

    \caption{MMCL challenges. We use a vision-language model architecture adapted from ViLT \cite{Kim2021Vilt} as an example to illustrate. \rev{(a) Graphical illustration of model components (Comp.) and challenges. (b) A corresponding summary table of the challenge-component relationships.} }
    
    \label{fig:MMCL challenges}
    \end{figure}

\renewcommand{\aboverulesep}{2pt}

\IEEEpubidadjcol

\hypertarget{mylink challenges}{
\textbf{Challenges of MMCL.} } In spite of the connection between conventional unimodal CL and MMCL, the challenges of MMCL extend beyond a simple stacking of CL\footnote{In this paper, we use terms {CL} and {MMCL} to respectively refer to unimodal and multimodal CL for simplicity, if no confusion is caused.} {methods on multimodal data.} 

\setcounter{challenge}{-1} %
\hypertarget{item: mmcl forgetting}{}\revsec{\begin{challenge}[Multimodal Catastrophic Forgetting]
    Stemming from unimodal CL, the primary and direct challenge in MMCL is multimodal catastrophic forgetting (\Cref{fig: non-tech illustration MMCL}).
    Straightforwardly applying unimodal CL methods has been demonstrated to yield suboptimal performance due to severe forgetting~\cite{Srinivasan2022CLiMB, He2024Continual, Greco2019Psycholinguistics}.
    Counterintuitively, despite the availability of more data, performance with multimodal inputs may be lower than with a single modality~\cite{Xu2024Continual, Sarfraz2024Unimodal}.
    Moreover, different modalities exhibit varying forgetting rates, further complicating this issue~\cite{He2024Continual}.
    \end{challenge}
}

\rev{
Beyond the severe forgetting phenomenon above, the multimodal nature of MMCL also introduces the following four challenges.
We detail these below and illustrate them in \Cref{fig:MMCL challenges}.
These challenges not only stand alone but may also exacerbate \challengeNumDesc{0}.
}

\begin{challenge}[Modality Imbalance]
\label{item:Modality imbalance} 
Modality imbalance refers to the uneven processing or representation of different modalities within a multimodal system, which manifests at both the \textit{data} and \textit{parameter} levels.
At the data level, the data availability of different modalities may significantly vary during the CL process, with extremely imbalanced cases such as the absence of certain modalities~\cite{Sun2021Multimodal}.
At the parameter level, the learning of different modality-specific components may converge at varying rates, leading to a holistic imbalanced learning process across all modalities~\cite{Chen2024Continual, Jin2024Calibrating}. 
This occurs because the modality with better performance may take a dominant position during optimization, whereas other modalities are under-optimized~\cite{Peng2022Balanced}.
\rev{After learning multimodal knowledge, a critical consequence of imbalance is forgetting rate differences among modalities. Some modalities exhibit greater degradation than others, failing to retain old knowledge (e.g., vision degrading faster than language)~\cite{He2024Continual, Li2025MMPrompt}. 
This imbalanced forgetting is amplified when modality-specific components encourage the model's disproportionate focus on features from its dominant modality, while other modalities are effectively suppressed~\cite{Li2025MMPrompt}.}
Therefore, MMCL models may suffer from performance degradation and, at times, may even perform worse than their unimodal counterparts~\cite{He2024Continual, Cheng2024VisionSensor}.

\end{challenge}

\begin{challenge}[Complex Modality Interaction]
\label{item:Modality interaction} 
{Modality interaction takes place in the model components where the representations of multimodal input information explicitly interact with one another.}
This interaction introduces unique challenges in MMCL, primarily manifesting in two interaction processes: \textit{modality alignment} and \textit{modality fusion}~\cite{Xu2023Multimodal}.
In modality alignment, features from different modalities of a data sample tend to diverge during continual learning, a phenomenon known as \textit{spatial disorder in MMCL}~\cite{Ni2023Continual}. This divergence may cause greater performance degradation, in contrast to the more robust nature of unimodal CL, such as in image-only settings~\cite{Ni2023Continual}. 
In modality fusion, a classical multimodal fusion approach used in the non-CL setting may perform worse in the MMCL setting, as different fusion techniques have varying effects on addressing the forgetting issue~\cite{Xu2024Continual,Cheng2024VisionSensor}.
In general, different modalities may exhibit inconsistent distributions and representations due to data heterogeneity~\cite{baltrusaitis2017multimodalmachinelearningsurvey,Peng2021Hierarchical}, and demonstrate different sensitivities to distribution shifts~\cite{Sarfraz2024Unimodal}, further complicating the alignment and fusion processes in MMCL.
In addition, the uncertainties in the modality interaction stage may contribute to model overfitting on downstream tasks and knowledge forgetting~\cite{Jha2024CLAP4CLIP}. 
Consequently, such complex modality interaction in MMCL highlights the necessity for specialized approaches to effectively incorporate multimodal data while maintaining CL capabilities.

\end{challenge}

\begin{challenge}[High Computational Costs]
\label{item:High computational costs}  
{The incorporation of multiple modalities in MMCL significantly increases computational costs at both the \textit{model} and \textit{task-specific} levels.}
At the model level, adding modalities inevitably increases the number of trainable parameters. 
Many MMCL methods utilize pre-trained multimodal models as their foundations. However, continuously fine-tuning these large-scale models in their entirety leads to heavy computational overhead~\cite{Yu2024Boosting,DAlessandro2023Multimodal}.
At the task-specific level, similarly, MMCL methods may lead to the consistent accumulation of task-specific trainable parameters, which can potentially exceed the number of parameters in the backbone model, thereby negating the original efficiency benefits of employing CL approaches~\cite{Jha2024CLAP4CLIP}.
These escalating computational demands pose strict requirements on the scalability of MMCL methods for practical deployment, especially given resource constraints.

\end{challenge}

\begin{challenge}[Degradation of Pre-trained Zero-shot Capability]
\label{item:Forgetting of the pre-trained knowledge} 
With advances in pre-trained models, MMCL methods can be armed with these powerful foundations. 
Consequently, these pre-trained multimodal models exhibit \textit{zero-shot} capability on unseen tasks~\cite{Zheng2024AntiForgetting, Zheng2023Preventing}, which distinguishes MMCL methods from traditional unimodal CL methods that usually train from scratch. 
However, during continuous fine-tuning of MMCL, some initial capabilities derived from pre-training foundations, such as performing zero-shot tasks, may diminish.
Such degradation risk may lead to severe performance decay on future tasks~\cite{Zheng2023Preventing}, known as \textit{negative forward transfer in MMCL}~\cite{Zheng2024AntiForgetting}. 
This phenomenon highlights that MMCL approaches must maintain the delicate balance between retaining pre-trained capabilities and adapting to new tasks.

\end{challenge}

\noindent \textbf{Contributions.} 
To address these challenges, researchers are increasingly focusing on MMCL methodologies. As detailed in Section~\ref{sec:Method}, our taxonomy categorizes the MMCL methods into four main approaches, i.e., regularization-based, architecture-based, replay-based, and prompt-based methods.
Given the increasing importance and research interest in MMCL, we present the \textit{first} comprehensive MMCL survey.
Beyond discussing the challenges faced in MMCL, this survey systematically details the basic formulations and settings (Section~\ref{sec:Setup}), reviews existing methodologies (Section~\ref{sec:Method}), summarizes relevant datasets and benchmarks (Section~\ref{sec:Benchmark}), 
\rev{provides a thorough discussion of the field (Section~\ref{sec: Discussion})},
and outlines promising future directions (Section~\ref{sec:Future Direction}). 
Our goal is not only to consolidate current MMCL advancements but also to inspire innovative research, thereby fostering the development of more effective and efficient MMCL approaches.
In summary, our survey makes the following key contributions:

\begin{enumerate}[leftmargin=*,label=(\arabic*), ref=(\arabic*)]
\item We present the first comprehensive survey on MMCL. We start by detailing essential MMCL
background knowledge, including the basic formulation, distinct MMCL scenarios, and widely used evaluation metrics.

\item In our structured taxonomy of MMCL methods, we categorize existing MMCL works into four categories with thorough subcategory explanations. For each category, we provide representative architecture illustrations and offer a detailed methodology review, highlighting key features and innovations accordingly. 

\item We summarize the current datasets and benchmarks to facilitate research and experiments. 
\rev{For this rapidly evolving field of MMCL, we provide an in-depth discussion of research trends, foundation models in MMCL, and the evolved stability-plasticity trade-off. }
We discuss promising future research directions, providing insights into potential areas for further investigation and development.
\end{enumerate}

\noindent\textbf{Connections with Other CL Surveys.} 
Several surveys are available mainly for general CL methodologies~\cite{Wang2024Comprehensive, DeLange2021Continual, Zhou2024ClassIncrementalLearning}.
\rev{There are also CL surveys focusing on the specific unimodal modality, such as computer vision~\cite{Masana2022ClassIncremental,qu2023recentadvancescontinuallearning} and natural language processing~\cite{Ke2022Continual, Wu2024Continual,Zheng2024Lifelong}.}
Additionally, with the advancing of pre-trained models and foundation language models, two works specifically review these developments~\cite{Zhou2024Continual,Yang2024Recent}.
There are CL surveys that specifically focus on other aspects like federated learning settings~\cite{Yang2024Federated} and knowledge distillation~\cite{Li2024Continual}, but they do not discuss MMCL methods.
Our work aims to present a comprehensive MMCL survey, addressing the lack of a dedicated survey in this area and filling this gap.

\newcommand*{\rulefiller}[2]{%
  \arrayrulecolor{lightblue} %
  \specialrule{#1}{0pt}{#2} %
  \arrayrulecolor{black} %
}

\begin{table}[tb]
  \caption{Notations and descriptions. \\The notations of $\mathcal{X}_t, p(\mathcal{X}_{t}), \mathcal{Y}_t \text{, and } \mathcal{T}$ are adopted from~\cite{Wang2024Comprehensive}.}
  \label{tab:Symbol}
  \centering
\rowcolors{1}{white}{lightblue}
  
  \scalebox{1}{
  \begin{tabular}{lp{6.5cm}}
      \toprule
      \textbf{Notation} & \textbf{Description} \\
      \midrule 
      \specialrule{6pt}{0pt}{-6pt} \rulefiller{6pt}{-4.5pt} $ t $ & Task-ID; $t\in \mathcal{T}=\{1,2, \cdots,T\}$; \newline
      $T\in \mathbb{N}$ represents the total number of tasks. \\
      $\mathcal{D}_t$ & The dataset of the $t$-th task; \\
      $\mathcal{D}$ & The entire dataset; \newline
      \mbox{$\mathcal{D}=\mathcal{D}_1\cup \mathcal{D}_2\cup \cdots\cup\mathcal{D}_T =\bigcup_{t=1}^{T} \mathcal{D}_t$}.  \\
      $\mathcal{X}_t$ & The input data of the $t$-th task.\\
      $\mathcal{X}$ & $\mathcal{X}=\bigcup_{t=1}^{T} \mathcal{X}_t$.\\
      $p(\mathcal{X}_{t})$ & The distribution of $\mathcal{X}_{t}$. \\
      $\mathcal{Y}_t$ & The data label of the $t$-th task. \\
      $\mathcal{Y}$ & $\mathcal{Y}=\bigcup_{t=1}^{T} \mathcal{Y}_t$. \\
      $\bm{\theta}$ & Trainable parameters. \\
      $\bm{\theta}^*_t$ & Optimal parameters after training on the $t$-th task.\\
      $\mathcal{M}$ & Episodic memory.\\
      $\mathcal{I}$ & The set of all input modalities present in $\mathcal{D}$; \newline 
      \mbox{$\mathcal{I}={\{1, 2, \cdots, I\}}$}; $I\in \mathbb{N}$ represents the total number of modalities; \newline
      $m \in \mathcal{I}$ represents the modality-ID that labels each modality. \\
      $\mathcal{I}_t$ & The set of input modalities present in $\mathcal{D}_t$; $\mathcal{I}_t\subseteq\mathcal{I}$. \\
      \rulefiller{1pt}{0pt} 
      \specialrule{1pt}{0pt}{-6pt}
  \end{tabular}
  }
\end{table}

\section{Preliminaries}
\label{sec:Setup}
In this section, we introduce the setup for MMCL, including notations, basic formulation, distinct learning scenarios, and widely used evaluation metrics.

\subsection{Notations}
\label{sec: Notations}
We use bold lowercase, bold uppercase, and calligraphy letters for vectors, matrices, and sets, respectively. We list the key notations in \Cref{tab:Symbol}.

\begin{figure*}[!t]
\hspace{-0.33cm} %
\captionsetup[subfigure]{justification=centering}
\subfloat[\small{Unimodal CL. \\The model is trained from scratch.}]{
    \label{fig: Illustration Unimodal CL}
    \includegraphics[width=0.32\linewidth,page=7]{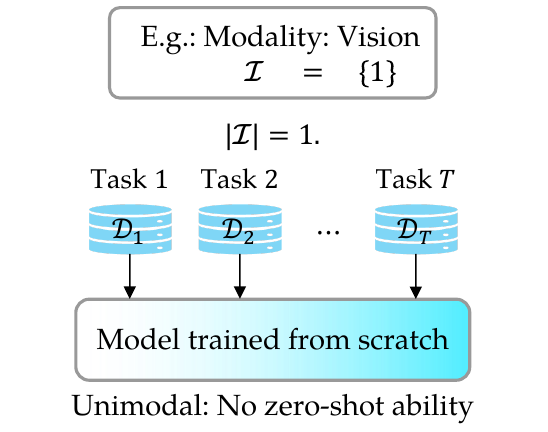}
}
\subfloat[\small{\rev{Multimodal CL. \\The model is trained from scratch.}}]{
    \label{fig: Illustration MMCL scratch}    
    \includegraphics[width=0.32\linewidth,page=8]{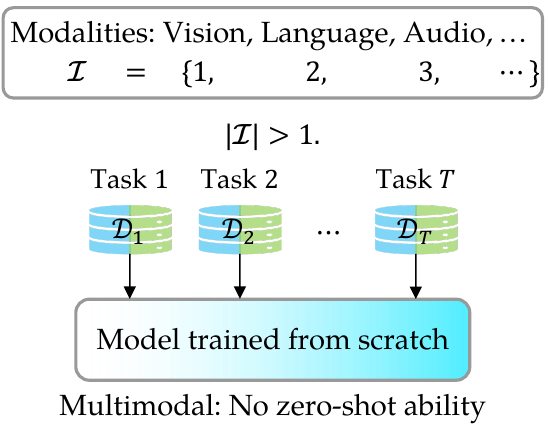}  
}
\subfloat[\small{\rev{Multimodal CL. The model is trained using a pre-trained MM backbone.}}]{
    \label{fig: Illustration MMCL pre-train}
    \includegraphics[width=0.32\linewidth,page=9]{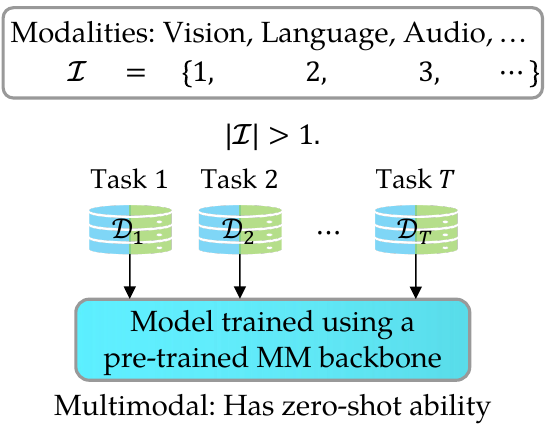}  
}
\caption{Illustrations of CL and MMCL. Notations are defined in \Cref{tab:Symbol}. }
\label{fig: Illustration of CL and MMCL}
\end{figure*}

\subsection{Basic Formulation}
In this section, we introduce the basic formulation of CL and MMCL.
\Cref{def:task sequence,def:CL} define task sequence and CL, respectively.
Then, we present the first formal definitions related to MMCL in \Cref{def:Modality-IDs and Set,def:Unimodal and Multimodal,def:modality properties,def:Subsequence,def:modality-dynamic,def:MMCL}.

\begin{definition}[Task Sequence]
\label{def:task sequence}
Let $\mathcal{X}_t$ and $\mathcal{Y}_t$ denote the input data and the data label of the $t$-th task, respectively.
The dataset of the $t$-th task, denoted as $\mathcal{D}_t$, is defined as:
\begin{equation}
    \mathcal{D}_t = \{(\mathbf{x}_{t,i}, y_{t,i}): i\in \mathbb{N}, 1\leq i\leq N_t\},
\end{equation}
where $\mathbf{x}_{t,i} \in \mathcal{X}_{t}$ and $y_{t,i} \in \mathcal{Y}_t$ are the $i$-th data, and $N_t$ is the number of samples of the $t$-th task. 
A \textbf{task sequence} $\revmath{\taskSeq}$ of size $T$ (where $T>1$ is required) is a sequence of tasks with their datasets in a certain order, defined as:
\begin{equation}
    \revmath{\taskSeq}\ =[\mathcal{D}_1, \mathcal{D}_2, \cdots, \mathcal{D}_T]. 
\end{equation}
We define the set of task-IDs as $\mathcal{T}=\{1,2, \cdots,T\}$.
\mbox{$\forall t\in \mathcal{T}, \revmath{\taskSeq}[t]=\mathcal{D}_t$}.
\end{definition}

\begin{definition}[Continual Learning (CL)]
\label{def:CL}
Given a task sequence $\revmath{\taskSeq}$ of size $T$, we consider the $t$-th task ($1<t\leq T$) as a new task so far.
\textbf{Continual learning} is the setting that, for each such task, the model is trained only on data $\mathcal{D}_t$ (or with very limited access to previous datasets $\{\mathcal{D}_1, \mathcal{D}_2, \cdots, \mathcal{D}_{t-1}\}$ in a more relaxed setting). 
The objective is to learn the new task while maintaining performance on old tasks to overcome catastrophic forgetting. 
Specifically, given an unseen test sample $\mathbf{x} \in \mathcal{X}$ from any trained tasks, the trained model $f: \mathcal{X} \to \mathcal{Y}$ should perform well in inferring the label $y = f(\mathbf{x}) \in \mathcal{Y}$~\cite{Wang2022Learning}.

\end{definition}

\begin{remark}
The performance of the model is evaluated using metrics described in Section~\ref{sec: Evaluation Metric}.
The difficulty of CL stems from the fact that datasets have dynamic distributions, i.e., \mbox{$\forall i,j \in \mathcal{T},\ i\neq j\Rightarrow\ p(\mathcal{X}_{i}) \neq p(\mathcal{X}_{j})$}~\cite{Wang2024Comprehensive}. 
\end{remark}

\begin{definition}[Modality-IDs and Set]
\label{def:Modality-IDs and Set}
Let $\mathcal{D}$ be the union of datasets of all tasks, defined as $\mathcal{D}=\bigcup_{t=1}^{T} \mathcal{D}_t$. 
\rev{
    Let $I$ be the total number of input modalities (e.g., vision, language, audio, etc.) present in $\mathcal{D}$. 
}
We define the \textbf{modality set} \mbox{$\mathcal{I}={\{1, 2, \cdots, I\}}$}.
Let $m \in \mathcal{I}$ represent the \textbf{modality-ID}, labeling each modality as a mathematical abstraction.
$\forall t \in \mathcal{T}$, let $\mathcal{I}_t$ be the \textbf{modality set} of $\mathcal{D}_t$, $\mathcal{I}_t\subseteq\mathcal{I}$.
\end{definition}

\begin{definition}[Unimodal and Multimodal] 
\label{def:Unimodal and Multimodal}
Given a task sequence $\revmath{\taskSeq}$ of size $T$, we say that
\begin{enumerate}[leftmargin=*,label=(\arabic*), ref=(\arabic*)]	
    \item $\revmath{\taskSeq}$ is \textbf{unimodal} if $|\mathcal{I}| = 1$; i.e., $\mathcal{D}$ contains one modality;
    \item $\revmath{\taskSeq}$ is \textbf{multimodal} if $|\mathcal{I}| > 1$; i.e.,  $\mathcal{D}$ contains more than one modality.
\end{enumerate}
\end{definition}

\begin{definition}[Modality-static, Modality-increasing, Modality-decreasing and Modality-switching]
\label{def:modality properties}
Given a task sequence $\revmath{\taskSeq}$ of size $T$, we say that
\begin{enumerate}[leftmargin=*,label=(\arabic*), ref=(\arabic*)]	
    \item $\revmath{\taskSeq}$ is \textbf{modality-static} if $\forall i,j \in \mathcal{T}, \mathcal{I}_i=\mathcal{I}_j$; i.e., all datasets have the same modality (or modalities);
    \item $\revmath{\taskSeq}$ is \textbf{modality-increasing} if $\forall 1<i\leq T, \mathcal{I}_{i-1} \subsetneq \mathcal{I}_i$; i.e., for each new task, it has more modalities compared to the previous task;
    \item $\revmath{\taskSeq}$ is \textbf{modality-decreasing} if $\forall 1<i\leq T, \mathcal{I}_{i-1} \supsetneq \mathcal{I}_i$; i.e., for each new task, it has fewer modalities compared to the previous task;
    \item $\revmath{\taskSeq}$ is \textbf{modality-switching} if $\forall 1<i\leq T, $ \mbox{$(\mathcal{I}_{i-1} \not\subseteq \mathcal{I}_i) \wedge (\mathcal{I}_{i-1} \not\supseteq \mathcal{I}_i)$}; i.e., two consecutive tasks have different modalities and do not have a subset relationship. For each new task, it switches modalities to involve different ones compared to the previous task.
\end{enumerate}

\end{definition}

\begin{definition}[Subsequence] 
\label{def:Subsequence}
Given a task sequence $\revmath{\taskSeq}$ of size $T$, a task sequence $\revmath{\taskSeq'}$ of size $T'$ is a \textbf{subsequence} of $\revmath{\taskSeq}$ 
if $\exists i\in \mathbb{N},$
\mbox{$\forall t\in \{1,2, \cdots,T'\},\revmath{\taskSeq'}[t]=\revmath{\taskSeq}[t+i]$} 
(or equivalently, 
\mbox{$\exists i\in \mathcal{T},$} 
\mbox{$ \revmath{\taskSeq'}=\revmath{\taskSeq}[i:i+T']$}
$=[\mathcal{D}_i, \mathcal{D}_{i+1}, \cdots, \mathcal{D}_{i+T'-1}]$).
\end{definition}

\begin{definition}[Modality-dynamic]
\label{def:modality-dynamic}
A multimodal task sequence $\revmath{\taskSeq}$ is \textbf{modality-dynamic} if it has a subsequence that is modality-increasing, modality-decreasing or modality-switching.
\end{definition}

\begin{definition}[Multimodal Continual Learning (MMCL)]
\label{def:MMCL}
Given a task sequence $\revmath{\taskSeq}$, 
\textbf{multimodal continual learning} is the setting where $\revmath{\taskSeq}$ is multimodal, and the model is trained under the CL setting.

\end{definition}

\begin{remark}
In addition to the challenge of catastrophic forgetting present in CL, MMCL introduces four challenges as described in Section~\hyperlink{mylink challenges}{\getrefnumber{sec: Introduction}}.
Moreover, when the task sequence is modality-dynamic, MMCL demonstrates increased flexibility but introduces greater complexity (e.g., ~\cite{Srinivasan2022CLiMB,Song2021Realworld,Sun2021Multimodal}). 
\end{remark}

Figure~\ref{fig: Illustration of CL and MMCL} provides graphical illustrations of CL and MMCL. Figure~\ref{fig: Illustration Unimodal CL} illustrates the case when in conventional CL, datasets have the same single modality, and the model is trained from scratch.
The model lacks the zero-shot ability. With multimodal datasets, MMCL methods may be trained either from scratch (\Cref{fig: Illustration MMCL scratch}) or using a pre-trained MM backbone (\Cref{fig: Illustration MMCL pre-train}). 
A model with a pre-trained MM backbone possesses zero-shot ability, i.e., to give zero-shot predictions on tasks. 
For example, the pre-trained CLIP model~\cite{Radford2021Learning} achieves zero-shot image classification accuracy of 88.5\% and 89.0\% on datasets of Food~\cite{Bossard2014Food101} and OxfordPet~\cite{Parkhi2012Cats}, respectively~\cite{Zheng2023Preventing}.
MMCL methods that use pre-trained MM backbones should address \Cref{item:Forgetting of the pre-trained knowledge} to preserve zero-shot capabilities throughout the learning process.

\newcommand{\mySizeCLScenarios}{5.13cm}
\newcommand{\mySizeMMCLScenarios}{6.8cm}

\newcommand{\mySizeCLScenariosHspace}{-0.28cm}
\newcommand{\mySizeMMCLScenariosHspace}{-0.33cm}

\begin{figure*}[!t]
\centering
\hspace{-0.33cm} %
\subfloat[\small{Class-incremental Learning}]{
    \label{fig: Illustration CIL}
    \includegraphics[height=\mySizeCLScenarios,page=11]{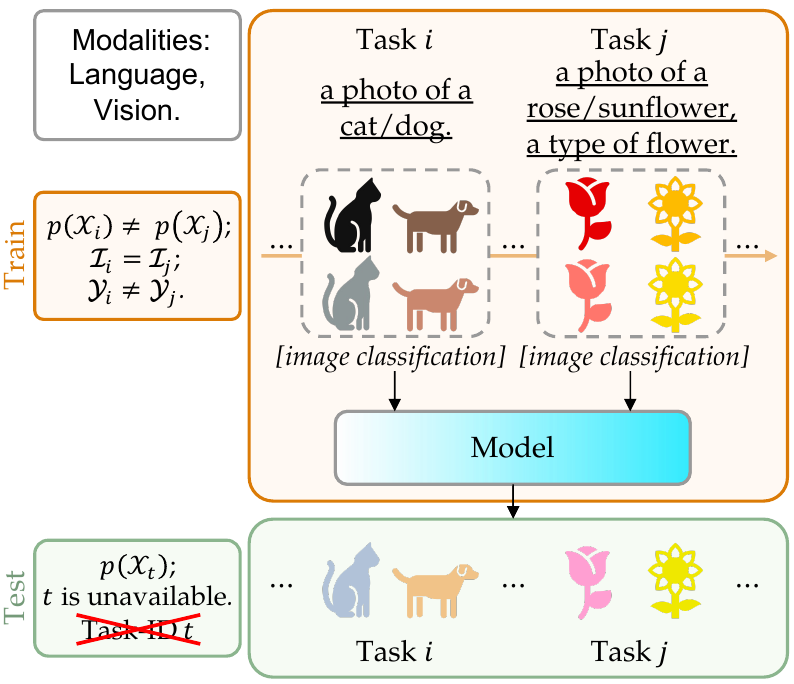}
    
}
\hspace{\mySizeCLScenariosHspace} %
\subfloat[\small{Domain-incremental Learning}]{
    \label{fig: Illustration DIL}    
    \includegraphics[height=\mySizeCLScenarios,page=12]{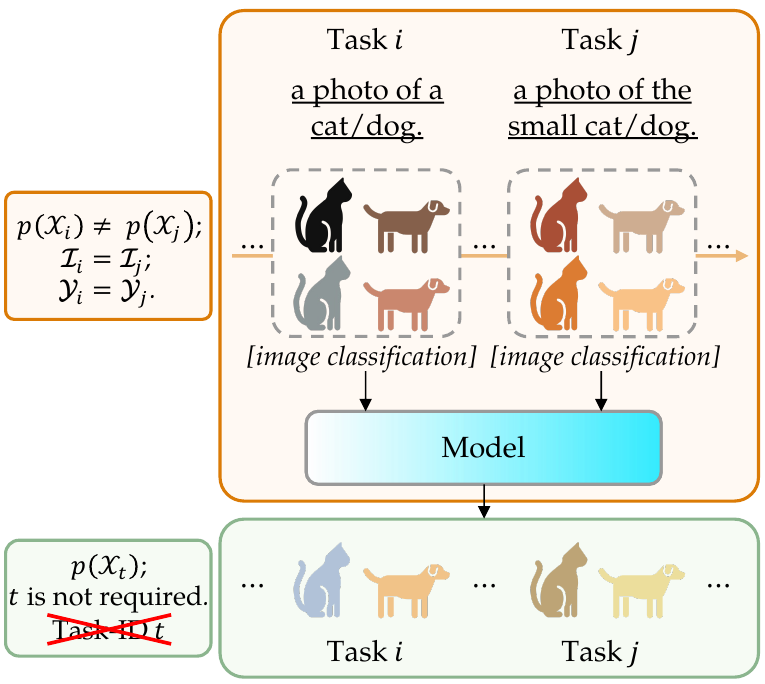}
}
\hspace{\mySizeCLScenariosHspace} %
\subfloat[\small{Task-incremental Learning}]{
    \label{fig: Illustration TIL}
    \includegraphics[height=\mySizeCLScenarios,page=13]{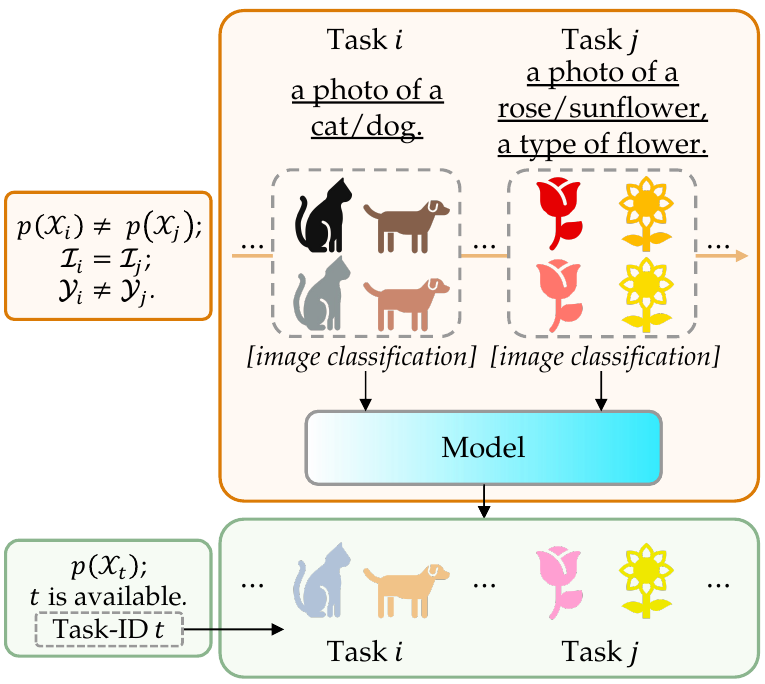}
}
\\
\hspace{-0.3cm} %
\subfloat[\small{\rev{Cross-Domain Incremental Learning (XDIL)}}]{
    \label{fig: Illustration GDIL}
    \includegraphics[height=\mySizeMMCLScenarios,page=14]{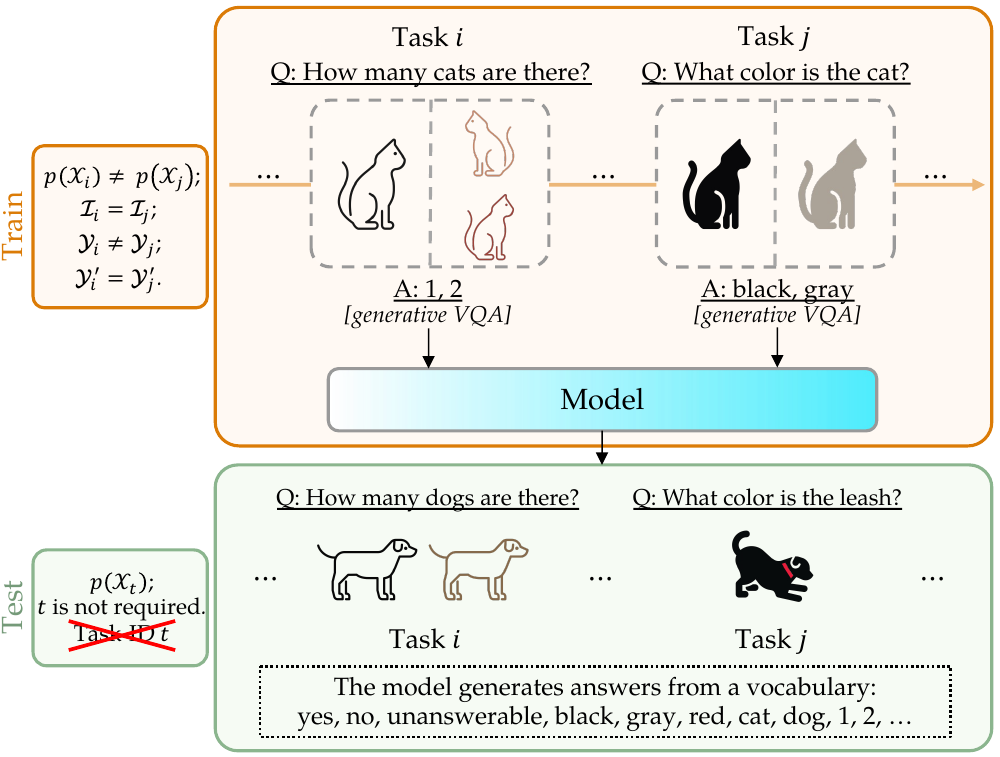}  
}
\hspace{\mySizeMMCLScenariosHspace} %
\subfloat[\small{Modality-dynamic Task-incremental Learning}]{
    \label{fig: Illustration MDTIL}
    \includegraphics[height=\mySizeMMCLScenarios,page=15]{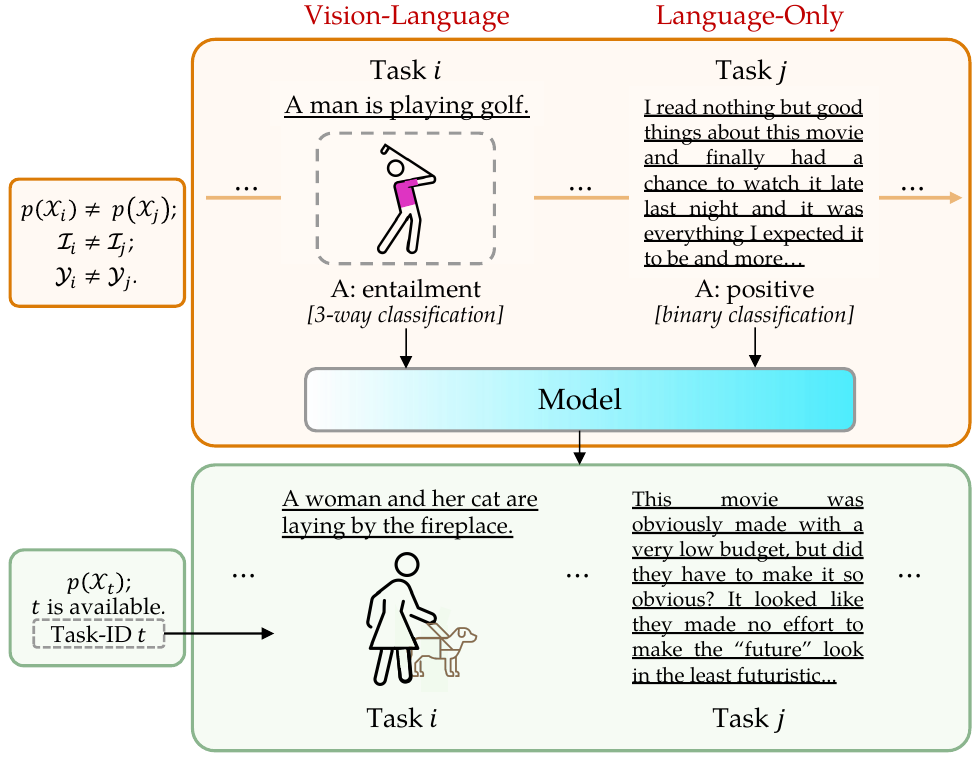}  
}
\caption{Illustrations of MMCL scenarios (defined in Section~\ref{sec: Continual Learning Scenarios}). Notations are defined in \Cref{tab:Symbol}. (a) Class-incremental Learning (CIL). (b) Domain-incremental Learning (DIL). (c) Task-incremental Learning (TIL). (d) \rev{Cross-Domain Incremental Learning (XDIL)}. (e) Modality-dynamic Task-incremental Learning (MDTIL). Figures \ref{fig: Illustration CIL}, \ref{fig: Illustration DIL} and \ref{fig: Illustration TIL} are partially adapted and redrawn based on~\cite{Masana2022ClassIncremental,Yang2024Recent}, with examples adapted based on MTIL~\cite{Zheng2023Preventing}, CIFAR10~\cite{Krizhevsky2009Learning} and Flowers~\cite{Nilsback2008Automated}. Examples in \Cref{fig: Illustration GDIL} are adapted based on VQAv2~\cite{Goyal2017Making}, VQACL~\cite{Zhang2023VQACL} and SGP~\cite{Lei2023Symbolic}. Examples in \Cref{fig: Illustration MDTIL} are adapted based on \climb\ \cite{Srinivasan2022CLiMB}, SNLI-VE~\cite{Xie2019Visual} and IMDb~\cite{Maas2011Learning}. }
\label{fig: Illustration of CL and MMCL scenarios}
\end{figure*}

\subsection{Multimodal Continual Learning Scenarios}
\label{sec: Continual Learning Scenarios}

In MMCL, the learning process varies in terms of modalities, data distribution, and task identity availability, resulting in five different MMCL scenarios.
We first introduce three scenarios that originate in conventional CL but can be inclusive in MMCL:

\begin{scenario}[Class-incremental Learning (CIL)]
For $i \neq j$, $\mathcal{D}_i$ and $\mathcal{D}_j$ have different input distributions and data label spaces, i.e., $p(\mathcal{X}_{i}) \neq p(\mathcal{X}_{j}) \wedge \mathcal{Y}_{i} \neq \mathcal{Y}_{j}$~\cite{Wang2024Comprehensive}. 
Task identities are not available in testing.
The model should be able to perform classification for all seen classes.
The model may need to infer the task-ID at test time to determine the possible classes of a test sample~\cite{vandeVen2022Three}. 
Note that in the conventional CIL setting, the data label spaces of tasks are disjoint, i.e., \mbox{$\forall i \neq j, \mathcal{Y}_{i} \cap \mathcal{Y}_{j} = \emptyset$}~\cite{Wang2024Comprehensive}; however, in a more generalized CIL setting, the data label spaces may overlap, i.e., \mbox{$\exists i \neq j, \mathcal{Y}_{i} \cap \mathcal{Y}_{j} \neq \emptyset$}~\cite{Sarfraz2024Unimodal}.
\end{scenario}

\begin{scenario}[Domain-incremental Learning (DIL)]
For \mbox{$i \neq j$}, $\mathcal{D}_i$ and $\mathcal{D}_j$ have different input distributions but the same label space, i.e., $p(\mathcal{X}_{i}) \neq p(\mathcal{X}_{j}) \wedge \mathcal{Y}_{i} = \mathcal{Y}_{j}$~\cite{Wang2024Comprehensive}.
Task identities are not required. 
Identifying the task is unnecessary for the model because of the same label space of all tasks~\cite{vandeVen2022Three}.

\end{scenario}

\begin{scenario}[Task-incremental Learning (TIL)]
For $i \neq j$, $\mathcal{D}_i$ and $\mathcal{D}_j$ have different input distributions and label spaces, i.e., $p(\mathcal{X}_{i}) \neq p(\mathcal{X}_{j}) \wedge \mathcal{Y}_{i} \neq \mathcal{Y}_{j}$~\cite{Wang2024Comprehensive}.
Task identities are available in testing.
The model needs to learn the tasks, and with the task-ID received at test time, it knows which task needs to be performed~\cite{vandeVen2022Three}. 

\end{scenario}

For MMCL, we introduce two new scenarios:

\begin{scenario}[\rev{Cross-Domain Incremental Learning (XDIL)}]
For $i \neq j$, $\mathcal{D}_i$ and $\mathcal{D}_j$ have different input distributions and label spaces, i.e., $p(\mathcal{X}_{i}) \neq p(\mathcal{X}_{j}) \wedge \mathcal{Y}_{i} \neq \mathcal{Y}_{j}$. 
Task identities are not required. 
\rev{This is a new scenario in MMCL, for tasks such as generative Visual Question Answering (VQA)~\cite{Zhang2023VQACL, Lei2023Symbolic} and cross-domain classification \cite{Xu2024Advancing, Yu2025Select} based on pre-trained vision-language models (VLMs).
The fundamental distinction between traditional CIL and XDIL lies in the dataset label spaces and model outputs.
In CIL, model predictions correspond to labels in the seen datasets.
However, in XDIL, the model outputs predictions from a larger set $\mathcal{Y}'$, e.g., the vocabulary set for VQA and the union set of all seen and unseen classes for classification. 
Labels of $\mathcal{D}_i$ are a subset of this predefined large set, $\mathcal{Y}_{i}\subset\mathcal{Y}'$. 
We view $\mathcal{Y}'_{i}:=\mathcal{Y}'$ as the actual label space. 
As such, the model's effective output label space remains consistent for all datasets, i.e., $\mathcal{Y}'_{i} = \mathcal{Y}'_{j}$, although input datasets have different label spaces. 
From the model's perspective, this consistency makes this scenario a domain incremental variant.
However, with a larger label space, it is more challenging and realistic than conventional CIL and DIL~\cite{Xu2024Advancing}.
We name it Cross-Domain Incremental Learning (XDIL) to emphasize the unique relationship between the input domains and the label space.}
\end{scenario}

\begin{figure*}[t]
  \centering
  \tikzset{
          my node/.style={
              draw,
              align=center,
              thin,
              text width=1.2cm, 
              rounded corners=3,
          },
          my leaf/.style={
              draw,
              align=left,
              thin,
              text width=8.5cm, 
              rounded corners=3,
          }
  }
  \forestset{
    every leaf node/.style={
      if n children=0{#1}{}
    },
    every tree node/.style={
      if n children=0{minimum width=1em}{#1}
    },
  }
  \begin{forest}
      nonleaf/.style={font=\bfseries\scriptsize},
       for tree={%
          every leaf node={my leaf, font=\scriptsize},
          every tree node={my node, font=\scriptsize, l sep-=4.5pt, l-=1.pt},
          anchor=west,
          inner sep=3pt,
          l sep=14pt, %
          s sep=5pt, %
          fit=tight,
          grow'=east,
          edge={ultra thin},
          parent anchor=east,
          child anchor=west,
          if n children=0{}{nonleaf}, 
          edge path={
              \noexpand\path [draw, \forestoption{edge}] (!u.parent anchor) -- +(7pt,0) |- (.child anchor)\forestoption{edge label};
          }, %
          if={isodd(n_children())}{
              for children={
                  if={equal(n,(n_children("!u")+1)/2)}{calign with current}{}
              }
          }{}
      }
      [Multimodal Continual Learning, draw=gray, fill=gray!15, text width=1.8cm, text=black
      [Regularization-based \\ ({\cref{sec: Continual Learning_Regularization-based}}), color=brightlavender, fill=brightlavender!15, text width=2.3cm, text=black
            [Modality Consistency Reg\\ ({\cref{sec:ER}}), \nodeParamsPattern{\nodeParamsPatternWidthSndLevel}{brightlavender}{brightlavender!15}
            [{
                TIR (\color{black}\citet{He2023Continual}\color{black}),
                SCD  (\color{black}\citet{Lao2023MultiDomain}\color{black}),  
                CS-VQLA  (\color{black}\citet{Bai2023Revisiting}\color{black}),
                SnD  (\color{black}\citet{Yu2025Select}\color{black})
                }, color=brightlavender, fill=brightlavender!15, text width=7.5cm, text=black]
            ],
            [Modality Relation Reg\\ ({\cref{sec:IR}}), \nodeParamsPattern{\nodeParamsPatternWidthSndLevel}{brightlavender}{brightlavender!15}
              [{ 
                ZSCL (\color{black}\citet{Zheng2023Preventing}\color{black}),  
                Mod-X (\color{black}\citet{Ni2023Continual}\color{black}),  
                CTP (\color{black}\citet{Zhu2023CTP}\color{black}),
                MSPT (\color{black}\citet{Chen2024Continual}\color{black}),
                MulKI (\color{black}\citet{Zhang2024MultiStage}\color{black})
              }, color=brightlavender, fill=brightlavender!15, text width=7.5cm, text=black],
            ],
          ]
      [Architecture-based \\ (\cref{sec: Continual Learning_Architecture-based}), color=lightgreen, fill=lightgreen!15, text width=2.3cm, text=black
            [Task-driven Architecture\\ ({\cref{sec:FA}}), \nodeParamsPattern{\nodeParamsPatternWidthSndLevel}{lightgreen}{lightgreen!15}
            [{
                RATT (\color{black}\citet{DelChiaro2020RATT}\color{black}) 
                MoE-Adapters4CL  (\color{black}\citet{Yu2024Boosting}\color{black}), 
                CLAP (\color{black}\citet{Jha2024CLAP4CLIP}\color{black}), 
                VLKD  (\color{black}\citet{Peng2021Hierarchical}\color{black}),
                EProj  (\color{black}\citet{He2023Continual}\color{black}),
                MSCGL  (\color{black}\citet{Cai2022Multimodal}\color{black}),
                CMR-MFN (\color{black}\citet{Wang2023Confusion}\color{black}),
                SS (\color{black}\citet{Ahrens2023Visually}\color{black}), 
                DIKI (\color{black}\citet{Tang2025Mind}\color{black})
                }, color=lightgreen, fill=lightgreen!15, text width=7.5cm, text=black]
            ],
            [Modality-driven Architecture\\ ({\cref{sec:DA}}), \nodeParamsPattern{\nodeParamsPatternWidthSndLevel}{lightgreen}{lightgreen!15}
              [{
                SCML (\color{black}\citet{Song2021Realworld}\color{black}),
                ODU (\color{black}\citet{Sun2021Multimodal}\color{black})
                }, color=lightgreen, fill=lightgreen!15, text width=7.5cm, text=black],
            ]
          ]
      [Replay-based \\ (\cref{sec: Continual Learning_Replay-based}), color=lightcoral, fill=lightcoral!15, text width=2.3cm, text=black
            [Natural Multimodal Replay\\ ({\cref{sec:DR}}), \nodeParamsPattern{\nodeParamsPatternWidthSndLevel}{lightcoral}{lightcoral!15}
              [{ 
              TAM-CL  (\color{black}\citet{Cai2023TaskAttentive}\color{black}),
              VQACL  (\color{black}\citet{Zhang2023VQACL}\color{black}),
              KDR  (\color{black}\citet{Yang2023Knowledge}\color{black}),
              SAMM  (\color{black}\citet{Sarfraz2024Unimodal}\color{black}),
              GMM  (\color{black}\citet{Cao2024Generative}\color{black})
              RAPF (\color{black}\citet{Huang2024ClassIncremental}\color{black}),
              RAIL (\color{black}\citet{Xu2024Advancing}\color{black})
              }, color=lightcoral, fill=lightcoral!15, text width=7.5cm, text=black],
            ],
            [Interactive Multimodal Replay\\ ({\cref{sec:PR}}), \nodeParamsPattern{\nodeParamsPatternWidthSndLevel}{lightcoral}{lightcoral!15}
              [{ 
              IncCLIP (\color{black}\citet{Yan2022Generative}\color{black}),
              SGP  (\color{black}\citet{Lei2023Symbolic}\color{black}),
              FGVIRs (\color{black}\citet{He2024Continual}\color{black}),
              AID (\color{black}\citet{Cheng2024VisionSensor}\color{black}),
              exFeCAM (\color{black}\citet{Kushawaha2024Continual}\color{black}),
              ConDU (\color{black}\citet{Gao2025Enhanced}\color{black})
              }, color=lightcoral, fill=lightcoral!15, text width=7.5cm, text=black],
            ]
        ]
        [Prompt-based\\ (\cref{sec: Continual Learning_Prompt-based}), \nodeParamsPattern{2.3cm}{harvestgold!70}{harvestgold!10}
              [Multimodal Prompt\\ ({\cref{sec:MP}}), \nodeParamsPattern{\nodeParamsPatternWidthSndLevel}{harvestgold!70}{harvestgold!10}
                [{ 
                  S-liPrompts (\color{black}\citet{wang2023sprompts}\color{black}),
                  TRIPLET (\color{black}\citet{Qian2023Decouple}\color{black}),
                  CPE-CLIP (\color{black}\citet{DAlessandro2023Multimodal}\color{black}),
                  HPC (\color{black}\citet{Jin2024Calibrating}\color{black}),
                  DPeCLIP (\color{black}\citet{Lu2024Boosting}\color{black}),
                  MM-Prompt  (\color{black}\citet{Li2025MMPrompt}\color{black})
                }, color=harvestgold!70, fill=harvestgold!10, text width=7.5cm, text=black],
              ],
              [Universal Prompt\\ ({\cref{sec:UP}}), \nodeParamsPattern{\nodeParamsPatternWidthSndLevel}{harvestgold!70}{harvestgold!10}
                [{ 
                  Fwd-Prompt (\color{black}\citet{Zheng2024AntiForgetting}\color{black}),
                  CoLeCLIP (\color{black}\citet{Li2025Coleclip}\color{black})
                  }, color=harvestgold!70, fill=harvestgold!10, text width=7.5cm, text=black],
              ]
        ]
      ]
  \end{forest}
  \caption{\rev{Taxonomy of multimodal continual learning (MMCL). We divide MMCL methods into four categories: Regularization-based (Section~\ref{sec: Continual Learning_Regularization-based}), Architecture-based (Section~\ref{sec: Continual Learning_Architecture-based}), Replay-based (Section~\ref{sec: Continual Learning_Replay-based}) and Prompt-based (Section~\ref{sec: Continual Learning_Prompt-based}).}}
  \label{fig: Taxonomy of Continual Learning}
  \end{figure*}
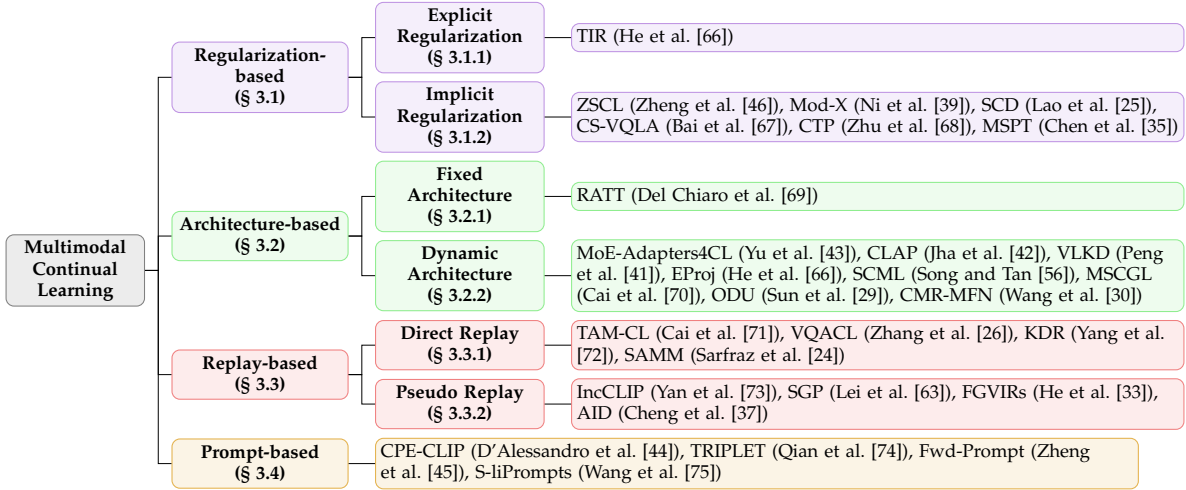

\begin{scenario}[Modality-dynamic Task-incremental Learning (MDTIL)]
In unimodal CL, the task sequence is naturally modality-static (\Cref{def:modality properties}) since there is only one modality. 
While in MMCL, on the one hand, if the task sequence is modality-static, it falls into one of the four scenarios described above.
On the other hand, the datasets may have different modalities, i.e., $\exists i,j \in \mathcal{T}, \mathcal{I}_i \neq \mathcal{I}_j$, and the task sequence is modality-dynamic (\Cref{def:modality-dynamic}, e.g., ~\cite{Srinivasan2022CLiMB, Song2021Realworld, Sun2021Multimodal}). 
For $i \neq j$, $\mathcal{D}_i$ and $\mathcal{D}_j$ have different input distributions and label spaces, i.e., $p(\mathcal{X}_{i}) \neq p(\mathcal{X}_{j}) \wedge \mathcal{Y}_{i} \neq \mathcal{Y}_{j}$.
\rev{Task identities are available in testing and are required to handle tasks with different input modalities~\cite{Srinivasan2022CLiMB, Song2021Realworld, Sun2021Multimodal}.}
We name this scenario as Modality-dynamic Task-incremental Learning (MDTIL).

\end{scenario}

We provide illustrations of all these five MMCL scenarios in~\Cref{fig: Illustration of CL and MMCL scenarios}. We use vision and language tasks as examples for illustration purposes, but MMCL scenarios can include tasks of various other modalities.

\subsection{Evaluation Metrics}
\label{sec: Evaluation Metric}
To evaluate the model performance in MMCL, various metrics are proposed.
In the single-task case, the performance evaluation metrics may vary depending on different task types.
For instance, these metrics may include accuracy for classification~\cite{Yu2024Boosting}, BLEU-4 for text generation~\cite{DelChiaro2020RATT}, Recall for retrieval tasks~\cite{Zhu2023CTP}, etc.
Based on these single-task evaluation metrics, we introduce common metrics for multiple tasks. 
Let $a_{t,i}$ $\in$ $[0,1]$ be the model performance on the test set of the $i$-th task, after the model is trained progressively from task $1$ to task $t$~\cite{Wang2024Comprehensive}.
\rev{Similarly, let $a_{t,i}^{(m)}$ be the performance using only modality $m$, if applicable.}
\begin{enumerate}[leftmargin=*,label=(\arabic*), ref=(\arabic*)]
\item \textit{Average Performance (A).} The average performance at the $t$-th task is defined as:
\begin{equation}
    \text{A}_t = \frac{1}{t} \sum_{i=1}^{t} a_{t,i}.
\end{equation}

\item \textit{Forgetting Measures (F)~\cite{Chaudhry2018Riemannian}.} 
Forgetting is quantified as the difference between the ``maximum'' knowledge and the current knowledge of a task during the continual learning process.
Let $f^t_i \in [-1,1]$ be the forgetting measure of the $i$-th task ($i < t$), after the model is progressively trained from task $1$ to task $t$:
\begin{equation}
    f^t_i = \max_{s\in\{1,\cdots,t-1\}} a_{s,i} - a_{t,i}, \forall i<t.
\end{equation}
The average forgetting at the $t$-th task is defined as: 
\begin{equation}
    \text{F}_t = \frac{1}{t-1}\sum_{i=1}^{t-1}f^t_i.
\end{equation}

\item \textit{Backward Transfer (BWT)~\cite{Lopez-Paz2017Gradient,Wang2024Comprehensive}. }
The difference \mbox{$a_{t,i} - a_{i,i}$} measures the influence of a task $t$ on a previous task $i$ ($i < t$). 
For the $t$-th task, backward transfer measures its average influence on the performance of all previous tasks. 
\begin{equation}
    \text{BWT}_t = \frac{1}{t-1}\sum_{i=1}^{t-1} a_{t,i} - a_{i,i}.
\end{equation}

\item \textit{Forward Transfer (FWT)~\cite{Lopez-Paz2017Gradient}.}
Let $\bar{b}_i$ be the performance of the $i$-th task with random initialization. 
The difference $a_{i,t} - \bar{b}_t$ measures the influence of a task $i$ on a future task $t$ ($i < t$). 
For the $t$-th task, forward transfer is defined as:
\begin{equation}
    \text{FWT}_t = \frac{1}{t-1}\sum_{i=2}^{t} a_{i-1,i} - \bar{b}_i.
\end{equation}

\hypertarget{mylink MMCL metrics}{}\rev{While traditional CL metrics provide general assessments, MMCL methods propose new metrics that offer targeted analysis relating to MMCL challenges:}
\item \textit{Zero-shot Transfer~\cite{Zheng2023Preventing}.} 
A recent MMCL method, ZSCL, proposes the ``Transfer'' metric to measure the level of preserved zero-shot ability (\rev{\challengeNum{4}}) on a task $t$ (\mbox{$t>1$}), after training on tasks before it. 
For the $t$-th task:
\begin{equation}
    \text{Transfer}_t = \frac{1}{t-1}\sum_{i=1}^{t-1} a_{i,t}.
\end{equation}

\revsec{\item \textit{Average Performance over Modalities (AM)~\cite{Kushawaha2024Continual}.} 
\label{metric: AM}
For multimodal methods that support training and inference for single modality input~\cite{Sarfraz2024Unimodal, Kushawaha2024Continual}, the AM metric provides a way to assess performance across individual modalities. Proposed by exFeCAM, it is a modality-oriented metric, unlike conventional CL metrics.
Let $\text{A}_t^{(m)}$ be the average performance on modality $m$ at the $t$-th task.
For the $t$-th task, AM is the average over all modalities $\mathcal{I}$, defined as: 
\begin{equation}
    \text{AM}_t = \frac{1}{I} \sum_{m \in \mathcal{I}} \text{A}_t^{(m)}.
\end{equation}
\item \textit{Average Difference (AD)~\cite{Li2025MMPrompt}.} 
\label{metric: AD}
To explicitly quantify modality imbalance (\challengeNum{1}), MM-Prompt proposes the AD metric. It measures the performance gap between the best and worst modalities.
For the $t$-th task:
\begin{equation}
    \text{AD}_t = \frac{1}{t} \sum_{i=1}^{t} \max_{m \in \mathcal{I}} a_{t,i}^{(m)}-\min_{m \in \mathcal{I}} a_{t,i}^{(m)}.
\end{equation}
Lower AD values indicate better balance among modalities, making the effectiveness of different methods addressing \challengeNum{1} comparable.}

\revsec{\item \textit{Average Merge Effectiveness (AME)~\cite{Li2025MMPrompt}.} 
\label{metric: AME}
MM-Prompt proposes the AME metric to measure the contribution of using multimodal data over the best unimodal data. 
For the $t$-th task:
 \begin{equation}
    \text{AME}_t = \frac{1}{t} \sum_{i=1}^{t} a_{t,i}-\max_{m \in \mathcal{I}} a_{t,i}^{(m)}.
\end{equation}
Higher AME values indicate a greater performance gain from multimodal data, reflecting the method's capability to resolve \challengeNumDesc{2}.}
\end{enumerate}

\section{Methodology}
\label{sec:Method}

\newcommand{\mySizeArchitectures}{5.7cm}

\begin{figure*}[tb]
\centering
\subfloat[\small{\rev{Regularization-based methods}}]{
            \label{fig:Regularization-based Architectures}  
            \includegraphics[height=\mySizeArchitectures,page=1]{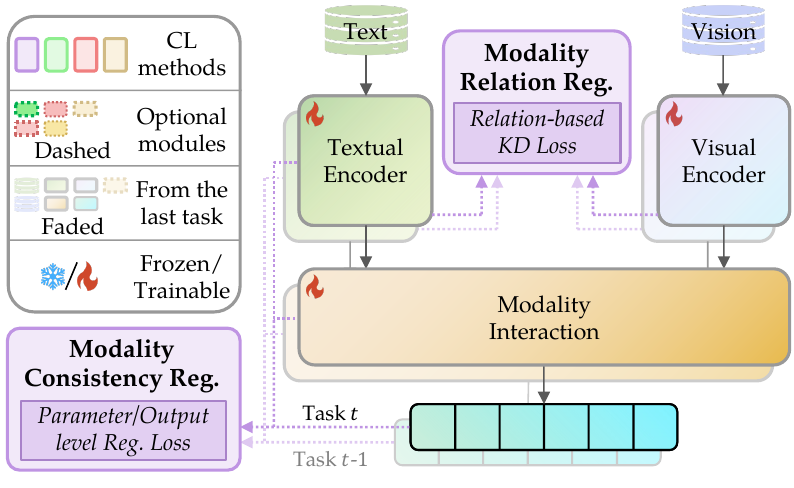}  
}
\subfloat[\small{\rev{Architecture-based methods}}]{
            \label{fig:Architecture-based Architectures}    
            \includegraphics[height=\mySizeArchitectures,page=2]{Architectures/MMCL3.pdf}  
}
\\
\subfloat[\small{\rev{Replay-based methods}}]{
            \label{fig:Replay-based Architectures}
            \includegraphics[height=\mySizeArchitectures,page=3]{Architectures/MMCL3.pdf}  
}
\subfloat[\small{\rev{Prompt-based methods}}]{
            \label{fig:Prompt-based Architectures}
            \includegraphics[height=\mySizeArchitectures,page=4]{Architectures/MMCL3.pdf}  
}
    \caption{Representative architectures for different categories of MMCL methods for vision and language. The base model architecture is adapted and redrawn based on ViLT \cite{Kim2021Vilt}. The prompt-based method architecture is partially adapted and redrawn based on TRIPLET \cite{Qian2023Decouple}. (a) \hyperref[sec: Continual Learning_Regularization-based]{Regularization-based}. (b) \hyperref[sec: Continual Learning_Architecture-based]{Architecture-based}. (c) \hyperref[sec: Continual Learning_Replay-based]{Replay-based}. (d) \hyperref[sec: Continual Learning_Prompt-based]{Prompt-based}.}
    \label{fig:Architectures}
\end{figure*}

In this section, we present a taxonomy of MMCL methods. Figure~\ref{fig: Taxonomy of Continual Learning} categorizes MMCL methods into four types, which we elaborate in the subsections below.
We summarize detailed properties of MMCL methods in \Cref{tab:summarizations} and the representative architectures of MMCL methods in \Cref{fig:Architectures}.
Note that \Cref{tab:summarizations} and \Cref{fig:Architectures} focus on methods of vision and language modalities, and methods of other modalities are summarized in \Cref{tab:summarization other modalities}.
\redsout{To ensure readability, we first introduce classical unimodal CL methods, as they are either the predecessors of various MMCL methods or are extensively compared in MMCL works.}

\subsection{Regularization-based Approach}
\label{sec: Continual Learning_Regularization-based}
Since the free movement of parameters in training causes catastrophic forgetting~\cite{Hassabis2017NeuroscienceInspired}, regularization-based methods are motivated to add constraints on the parameters.
Let $\mathcal{L}_{single,t}$ be the loss in the single task setting when the model is learning the $t$-th task.
\rev{The continual loss is then defined as $\mathcal{L}_{t}=\mathcal{L}_{single,t}+ \lambda \mathcal{L}_{R,t} $, where $\mathcal{L}_{R,t}$ is the regularization term and the hyperparameter $\lambda$ balances plasticity on the new task with stability for old tasks.}
\rev{Based on the purpose of regularization for preserving multimodal information, regularization-based methods are divided into two sub-directions: \textbf{\textit{modality consistency regularization}} and \textbf{\textit{modality relation regularization}}.
We summarize the representative architectures of the above two subcategories in \Cref{fig:Regularization-based Architectures}.}

\subsubsection{\rev{\textbf{Modality Consistency Regularization}}}
\label{sec:ER}
\revsec{Modality consistency regularization aims to maintain the consistency of learned modality-specific representations or fused information of previous tasks, while learning new tasks. This category primarily aims to directly address \textbf{\challengeNumDesc{0}}.
One approach to preserve consistency is to regularize at the \textit{parameter-level}, by directly assigning importance to parameters and penalizing them differently when they deviate from the previously found solution.
}
\rev{$\mathcal{L}_{R,t}$} can be formulated as follows~\cite{Kirkpatrick2017Overcoming}:
\begin{align}
\rev{\mathcal{L}_{R,t}}=\sum_i b_i\left(\theta_i-\theta_{t-1, i}^*\right)^2 ,
\label{eq:reg explicit}
\end{align}
where $\theta_i$ and $\theta^*_{t-1, i}$ denote the $i$-th element in $\bm{\theta}$ and $\bm{\theta}^*_t$ respectively, and $b_i$ indicates the corresponding importance.

\rev{
Instead of assigning weights to individual parameters, another approach keeps modality information consistent by minimizing changes in the model's output on past knowledge, i.e., regularizing at the \textit{output-level}.
Specifically, this output may be \textit{modality-specific features} from their respective encoders, or fused representations and model logits, which contain \textit{multimodal fused information}.
Unlike parameter-level regularization,
output-level regularization offers greater flexibility in the design of the constraint.
These methods typically employ \textit{logits-based} and \textit{feature-based} knowledge distillation (KD)~\cite{Hinton2015Distilling}, which matches the output between a teacher model (the previous-task trained model) and a student model (the current model)~\cite{Wang2024Comprehensive}.
Therefore, KD encourages the student model to learn logits and feature distribution similar to those of the teacher model~\cite{Huang2017What, Gou2021Knowledge}. }
\rev{$\mathcal{L}_{R,t}$} incorporates KD and can be formulated as follows~\cite{Li2017Learning,Bai2023Revisiting}:
\begin{align}
\revmath{\mathcal{L}_{R,t}} &=\mathcal{L}_{KD}(\boldsymbol{y}_{t-1},\boldsymbol{y}_t) \notag \\
&=
\begin{cases}
-\sum_{i} y_{t-1,i}\log y_{t,i} & \text{cross-entropy loss} \\
\|\boldsymbol{y}_{t-1}-\boldsymbol{y}_t\|_2^2   & \text{L2 loss},
\end{cases}
\label{eq:reg implicit}
\end{align}
where $\boldsymbol{y}_{t-1}$ and $\boldsymbol{y}_t$ are the outputs of one data sample from the model before and after training on the $t$-th task.
$y_{t-1,i}$ and $y_{t,i}$ denote the $i$-th element in $\boldsymbol{y}_{t-1}$ and $\boldsymbol{y}_t$, respectively.

\redsout{\textbf{Representative Unimodal Models.}
EWC~\cite{Kirkpatrick2017Overcoming} utilizes the diagonal of the Fisher information matrix as the term $b_i$ in Equation~(\mbox{\ref{eq:reg explicit}}) and accumulates multiple regularization terms for the previously found solutions of previous tasks.
This method restricts parameter changes that are crucial for previous tasks, while allowing greater flexibility for less significant parameters.
EWC is used extensively in unimodal and multimodal works for performance comparison because it is effective and model-agnostic. Several follow-up works like EWC\textsuperscript{H}~\cite{Huszar2018Quadratic} and online EWC~\cite{Schwarz2018Progress} have been proposed to further enhance the efficacy and efficiency of EWC by employing a single regularization term instead of multiple terms.}

\rev{\mmclMethods}
\rev{In the MMCL setting, methods propose modality consistency regularization algorithms built upon multimodal base models.
For example, SnD~\cite{Yu2025Select} utilizes CLIP~\cite{Radford2021Learning},
while TIR~\cite{He2023Continual} leverages Multimodal Large Language Models (MLLM) such as BLIP2~\cite{li2023blip2} and InstructBLIP~\cite{dai2023instructblip}.
These MMCL methods utilize regularization techniques at either the parameter-level or the output-level.
Therefore, we present methods based on the above two levels for the following explanations.}

\hypertarget{mylink TIR}{}\rev{At the \textit{parameter-level}, TIR~\cite{He2023Continual} measures parameter importance using existing methods such as EWC~\cite{Kirkpatrick2017Overcoming}.}
It calculates task similarity scores between new and old tasks to obtain adaptive weights for regularization, facilitating long-term CL.

\rev{At the \textit{output-level}, CS-VQLA~\cite{Bai2023Revisiting} and SCD~\cite{Lao2023MultiDomain} employ both \textit{logits-based} and \textit{feature-based KD} techniques.}
CS-VQLA proposes rigidity-plasticity-aware distillation (logits level) to deal with non-overlapping and overlapping classes separately in the CIL scenario. Moreover, it introduces self-calibrated heterogeneous distillation (feature level) to minimize the distance between the self-calibrated feature map and the old feature map.
SCD~\cite{Lao2023MultiDomain} proposes to transfer domain knowledge through self-critical distillation at both the logits and feature levels. It defines instance-relevant and domain-relevant knowledge based on teacher model prediction and proposes a self-critical temperature to adjust the knowledge transfer.
\rev{SnD~\cite{Yu2025Select} selectively employs feature-based KD with the most recent fine-tuned and the original pre-trained models as dual teachers, preserving both knowledge from past tasks and zero-shot capabilities (\challengeNum{4}).}

\renewcommand{\aboverulesep}{0pt}
\newcommand{\mySizeCA}{0.2cm}
\begin{table*}[t]
\centering
    \caption{\rev{A summary of MMCL methods for vision and language. }}
    \label{tab:summarizations}
\rowcolors{1}{white}{lightblue}
\resizebox{\linewidth}{!}{
\begin{NiceTabular}{lll|ccccc|l|l|ccc|>{\centering}m{\mySizeCA}m{\mySizeCA}m{\mySizeCA}m{\mySizeCA}|c}
\CodeBefore
\rowcolors{3}{blue!10}{}[cols=3-18,restart]
\Body
\toprule
 & & \multirow{2.5}{*}{\textbf{Method}} & \multicolumn{5}{c|}{\textbf{MMCL Scenario}} & \multirow{2.5}{*}{\textbf{Task}} & \multirow{2.5}{*}{\textbf{MM Backbone}} & \multicolumn{3}{c|}{\textbf{CL-}} & \multicolumn{4}{c|}{\textbf{CA}} & \multirow{2.5}{*}{\textbf{Code}}\\
 \cmidrule{4-8} 
 \cmidrule{11-17} 
  & & & CIL & DIL & XDIL & TIL & MDTIL & & & L & V & MI  & \challengehyperref{item:Modality imbalance} & \challengehyperref{item:Modality interaction} & \challengehyperref{item:High computational costs} & \challengehyperref{item:Forgetting of the pre-trained knowledge}  &   \\ 
  
\midrule

\multirow{9.5}{*}{\rotatebox{90}{Regularization}}  
& \multirow{4}{*}{CR} 
& TIR \cite{He2023Continual} & & & \cmark &  &  & GEN & BLIP2~\cite{li2023blip2}, InstructBLIP~\cite{dai2023instructblip} & \cmark &  & \cmark  &  &  &  &   & -  \\
& & SCD \cite{Lao2023MultiDomain} & & & & \cmark &  & CLS & ViLT \cite{Kim2021Vilt} & \cmark & \cmark & \cmark    &  &  & &   & -              \\
& &  CS-VQLA \cite{Bai2023Revisiting} & \cmark & & & &  & CLS & VisualBERT~\cite{Li2019VisualBERT} & \cmark & \cmark & \cmark &   &  &  &    & \href{https://github.com/longbai1006/CS-VQLA}{Link} \\
& &  SnD \cite{Yu2025Select} & & & \cmark & \cmark &  & CLS & CLIP \cite{Radford2021Learning} &  & \cmark &   &   &  &  &  \challengeMark   & \href{https://chuyu.org/research/snd/}{Link} \\
\cmidrule{2-18} 
& \multirow{5}{*}{RR} 
&  ZSCL \cite{Zheng2023Preventing} & \cmark & & & \cmark &  & CLS & CLIP & \cmark & \cmark &  &   & \challengeMark &  &   \challengeMark  & \href{https://github.com/Thunderbeee/ZSCL}{Link} \\
& &  Mod-X \cite{Ni2023Continual} & \cmark & & & & & RET & CLIP & \cmark & \cmark &  &   &  \challengeMark & &  & -              \\
& & CTP \cite{Zhu2023CTP} & \cmark & & & &  & RET & - & \cmark & \cmark & \cmark &      &  \challengeMark &  &  & \href{https://github.com/KevinLight831/CTP}{Link} \\
& &  MSPT \cite{Chen2024Continual} &  \cmark  &  &  &  &  & CLS & - & \cmark  & \cmark & \cmark  &   \challengeMark  &  \challengeMark  &  &  & \href{https://github.com/zjunlp/ContinueMKGC}{Link} \\
& &  MulKI \cite{Zhang2024MultiStage} &  \cmark   &  &  & \cmark &  & CLS & CLIP & \cmark  & \cmark &   &   &  \challengeMark  &  & \challengeMark & - \\

\midrule
                                  
\multirow{9.5}{*}{\rotatebox{90}{Architecture}} 
& \multirow{8}{*}{TA} 
& RATT \cite{DelChiaro2020RATT} & & & & \cmark &  & GEN & - & \cmark & \cmark & \cmark &    &  &  &  & \href{https://github.com/delchiaro/RATT}{Link}             \\
&  &  MoE-Adapters4CL \cite{Yu2024Boosting} & \cmark & & & \cmark & & CLS & CLIP\textsuperscript{\starMark}    & \cmark &    \cmark  &     &  &  &  \challengeMark  & \challengeMark  & \href{https://github.com/JiazuoYu/MoE-Adapters4CL}{Link}       \\
& &  DIKI \cite{Tang2025Mind} & \cmark   & & & &  & CLS & CLIP\textsuperscript{\starMark} & \cmark & \cmark &  &   &    & \challengeMark &  \challengeMark  & \href{https://github.com/lloongx/DIKI}{Link}       \\
& & CLAP \cite{Jha2024CLAP4CLIP} & \cmark & & & & & CLS & CLIP &  \cmark &      & \cmark  &    &  \challengeMark  &  & \challengeMark  & \href{https://github.com/srvCodes/clap4clip}{Link}       \\
& &  VLKD \cite{Peng2021Hierarchical} & & & & \cmark &  & RET & - & \cmark &   \cmark   & \cmark  &  &  \challengeMark  &  &   & -             \\
& & EProj \cite{He2023Continual} & & & \cmark &  &  & GEN & BLIP2, InstructBLIP &  &  & \cmark &    &  &  \challengeMark  &  \challengeMark  & - \\
& &  MSCGL \cite{Cai2022Multimodal} &  \cmark & & & &  & CLS & - & \cmark & \cmark & \cmark &   &  \challengeMark  &  &   & - \\
& &  SS \cite{Ahrens2023Visually} &   & & & \cmark &  & RET & - &  &  & \cmark &   &  \challengeMark  &  &    & \href{https://github.com/ky-ah/selective-lilac}{Link}       \\
\cmidrule{2-18} 
& \multirow{1}{*}{MA} 
&  SCML \cite{Song2021Realworld} & & & & &  \cmark & CLS, RET & - & \cmark & \cmark & \cmark & \challengeMark  &  \challengeMark  &  &   & - \\

\midrule
\multirow{9.5}{*}{\rotatebox{90}{Replay}} 
& \multirow{6}{*}{NR} 
&  TAM-CL \cite{Cai2023TaskAttentive} & & & & \cmark & & CLS & - & \cmark & \cmark & \cmark &  &  &  &   & \href{https://github.com/YuliangCai2022/TAM-CL}{Link}  \\
& &   VQACL \cite{Zhang2023VQACL} & & & \cmark & & & GEN & - & \cmark & \cmark & \cmark &  &  &  &    & \href{https://github.com/zhangxi1997/VQACL}{Link} \\
& &  KDR \cite{Yang2023Knowledge} & & \cmark & & & & RET & - & \cmark & \cmark & \cmark &    &  \challengeMark  &  &  & - \\
& &  GMM \cite{Cao2024Generative} & & & \cmark & & & CLS & BLIP2 &  & \cmark &  &  &  &  \challengeMark  &  \challengeMark  & \href{https://github.com/DoubleClass/GMM}{Link}  \\
& &  RAPF \cite{Huang2024ClassIncremental} & \cmark & &  & & & CLS & CLIP &   & \cmark &   &  &  &  \challengeMark  & \challengeMark  & \href{https://github.com/linlany/RAPF}{Link} \\
& &  RAIL \cite{Xu2024Advancing} &   &  &  \cmark &  &  & CLS & CLIP &   & \cmark &   &   &  &  \challengeMark  &  \challengeMark   & \href{https://github.com/linghan1997/Regression-based-Analytic-Incremental-Learning}{Link} \\

\cmidrule{2-18}
& \multirow{3}{*}{IR}
&   IncCLIP \cite{Yan2022Generative} & \cmark & \cmark & & & & CLS, RET & CLIP & \cmark & \cmark & \cmark &  &  \challengeMark  &  &  & - \\
& &  SGP \cite{Lei2023Symbolic} & & & \cmark & & & GEN & - & \cmark & \cmark & \cmark &  & \challengeMark &  &    & \href{https://github.com/showlab/CLVQA}{Link} \\
& &  ConDU \cite{Gao2025Enhanced} &   & & \cmark & \cmark  &  & CLS & CLIP & \cmark & \cmark &  &   &    &  &  \challengeMark  & -     \\

\midrule

\multirow{8.5}{*}{\rotatebox{90}{Prompt}} 
& \multirow{6}{*}{MP} &  CPE-CLIP \cite{DAlessandro2023Multimodal} & \cmark & & & & & CLS & CLIP\textsuperscript{\starMark}  & \cmark & \cmark &   &  & \challengeMark \ &  \challengeMark  &  \challengeMark & \href{https://github.com/neuraptic/cpe-clip}{Link} \\
& & TRIPLET \cite{Qian2023Decouple} & & & & \cmark & & CLS & ALBEF~\cite{Li2021Align}, FLAVA~\cite{Singh2022Flava}\textsuperscript{\starMark}  & \cmark & \cmark & \cmark &   &  \challengeMark  &  \challengeMark  & \challengeMark  & - \\
& & S-liPrompts \cite{wang2023sprompts} & & \cmark & & &  & CLS & CLIP\textsuperscript{\starMark}  & \cmark & \cmark &  &  &  &  \challengeMark  &  \challengeMark  & \href{https://github.com/iamwangyabin/S-Prompts}{Link} \\
&   &  HPC \cite{Jin2024Calibrating} & \cmark & & & & & RET & CLIP, GLIP~\cite{Li2022Grounded}\textsuperscript{\starMark}   & \cmark  & \cmark  &  &  & \challengeMark &  \challengeMark  &  \challengeMark  & - \\
&   &  DPeCLIP \cite{Lu2024Boosting} & & & \cmark &  & & CLS & CLIP\textsuperscript{\starMark}   & \cmark  & \cmark  & \cmark &  &  &  \challengeMark  &  \challengeMark  & \href{https://github.com/DeepMed-Lab-ECNU/DPeCLIP}{Link} \\
&   &  MM-Prompt \cite{Li2025MMPrompt} & & & \cmark & & & GEN & LXMERT~\cite{Tan2019LXMERT} + T5~\cite{Raffel2020Exploring}\textsuperscript{\starMark}   & \cmark  & \cmark  & \cmark & \challengeMark  & \challengeMark  &  \challengeMark  &  \challengeMark  & \href{https://github.com/xli04/CVQA}{Link} \\
\cmidrule{2-18}
& \multirow{2}{*}{UP} &  Fwd-Prompt \cite{Zheng2024AntiForgetting} & & & \cmark & & & GEN & BLIP2, InstructBLIP\textsuperscript{\starMark}  & \cmark & \cmark & \cmark  &  &  &  \challengeMark  &  \challengeMark  & - \\
&  &  CoLeCLIP \cite{Li2025Coleclip} & & & \cmark & \cmark & & CLS & CLIP\textsuperscript{\starMark}  & \cmark & \cmark &   &  &  &  \challengeMark  &  \challengeMark  & \href{https://github.com/YukunLi99/CoLeCLIP}{Link}  \\

\bottomrule
\end{NiceTabular}}

\begin{flushleft}
\smallskip
``MMCL Scenario'': defined in Section~\ref{sec: Continual Learning Scenarios}; ``MM Backbone'': the MM backbone of the MMCL methods, and \starMark \ indicates that the model uses parameter-efficient fine-tuning (PEFT) strategies; ``Task'': \textit{CLS} means classification for input modality (or modalities), \textit{RET} means image-text retrieval, and \textit{GEN} means text generation; ``CL-V/L/MI (Vision/Language/Modality Interaction)'': \cmark \ indicates that the model continually learns vision information, language information, and modality interaction, respectively; 
\revsec{``CA (Challenges Addressed)'': \challengeMark \ indicates challenges 1 to 4 described in Section~\texorpdfstring{\protect\hyperlink{mylink challenges}{\getrefnumber{sec: Introduction}}{}}\ \ that the method has addressed. Note that because all methods aim to address challenge 0, we omit it in the table to enhance clarity and emphasize efforts towards new MMCL challenges;}
``Code'': the open-source implementation. ``-'' represents non-existence.
\end{flushleft}

\end{table*}

\renewcommand{\aboverulesep}{2pt}

\subsubsection{\rev{\textbf{Modality Relation Regularization}}}
\label{sec:IR}
\rev{With paired multimodal data samples, such as image-text pairs, modality relation regularization aims to preserve the learned pairwise relations between modalities from previous tasks, thereby mitigating the forgetting of cross-modal alignments. This category mainly addresses \textbf{\challengeNumDesc{2}}.
Methods in this category typically utilize \textit{relation-based} KD~\cite{Gou2021Knowledge}, which guides the current model to mimic the inter-modality relations captured by the previous-task trained model.
Let $\mathcal{F}^{(m_1)}$ and $\mathcal{F}^{(m_2)}$ be the feature sets of two modalities.
$\mathcal{L}_{R,t}$ incorporates relation-based KD and can be formulated as follows~\cite{Gou2021Knowledge}:
\begin{align}
\mathcal{L}_{R,t}=\mathcal{L}_{R} \left( \Psi (\mathcal{F}^{(m_1)}_{t-1}, \mathcal{F}^{(m_2)}_{t-1}), \Psi (\mathcal{F}^{(m_1)}_{t}, \mathcal{F}^{(m_2)}_{t}) \right),
\label{eq: reg relation}
\end{align}
where $\Psi(\cdot,\cdot)$ is the relational function. Its output represents the relational knowledge, which can be a matrix (for many-to-many relations) or a vector (for one-to-many relations).
$\mathcal{L}_{R}(\cdot)$ is the distillation loss function, such as the Frobenius norm~\cite{Zhang2024MultiStage} and cross-entropy loss~\cite{Zheng2023Preventing}.}

\redsout{\textbf{Representative Unimodal Models.}
LwF~\cite{Li2017Learning} is a classical regularization-based CL work that incorporates the KD design.
It calculates the output of old tasks using new data before training on a new task.
During the learning process for the new task, the model minimizes the changes in the outputs of previous tasks using the KD loss.
This strategy avoids the need for explicit storage or reuse of data from previous tasks.
LwF is widely employed in both unimodal and multimodal studies for performance comparisons.}

\renewcommand{\aboverulesep}{0pt}

\begin{table*}[t]
\centering
\caption{\rev{A summary of MMCL methods focusing on modalities other than vision and language.}}

\label{tab:summarization other modalities}

\resizebox{\linewidth}{!}{
\begin{NiceTabular}{lll|ccccc|cccccc|c|m{\mySizeCA}m{\mySizeCA}m{\mySizeCA}m{\mySizeCA}|c}
\CodeBefore
\rowcolors{3}{blue!10}{}[cols=3-20,restart]
\Body
\toprule
& & \multirow{2.5}{*}{\textbf{Method}} & \multicolumn{5}{c|}{\textbf{MMCL Scenario}} & \multicolumn{6}{c|}{\textbf{Modality}} & \multirow{2.5}{*}{\textbf{Task}} & \multicolumn{4}{c|}{\textbf{CA}} & \multirow{2.5}{*}{\textbf{Code}}\\
\cmidrule{4-14} 
\cmidrule{16-19} 
& & & CIL & DIL & XDIL & TIL & MDTIL &  Vision & Language & Audio & Tactility & Acceleration & Gyroscope & & \challengehyperref{item:Modality imbalance} & \challengehyperref{item:Modality interaction} & \challengehyperref{item:High computational costs} & \challengehyperref{item:Forgetting of the pre-trained knowledge}  \\ 
 \midrule
\multirow{2.5}{*}{\rotatebox{90}{\parbox[c]{0.73cm}{Archi-tecture}}} 
& \multirow{1}{*}{TA}  &  CMR-MFN \cite{Wang2023Confusion} & \cmark  &  &  &  &  & \cmark &  &  &  & \cmark  & \cmark & CLS &  &  \challengeMark  &  &  & \href{https://github.com/Hanna-W/CMR-MFN}{Link} \\ %
\cmidrule{2-20} 
& \multirow{1}{*}{MA} &  ODU \cite{Sun2021Multimodal} &   & &  &  & \cmark &  \cmark &  & \cmark & \cmark &  &  & CLS &  \challengeMark  &  \challengeMark  &  & & - \\

\midrule
\multirow{4.5}{*}{\rotatebox{90}{Replay}}
& \multirow{1}{*}{NR} & SAMM~\cite{Sarfraz2024Unimodal} & \cmark  & \cmark &  &  &  & \cmark &  & \cmark &  &  &  & CLS &  &  \challengeMark  &  &  & \href{https://github.com/NeurAI-Lab/MultiModal-CL}{Link}  \\
\cmidrule{2-20} 
& \multirow{3}{*}{IR} &  AID \cite{Cheng2024VisionSensor} & \cmark  &  &  &  &  & \cmark &  &  &  & \cmark  & \cmark & CLS &  \challengeMark  &  \challengeMark  &  &  & - \\
&  & FGVIRs~\cite{He2024Continual} & \cmark  &  &  &  &  & \cmark &  &  &  & \cmark  & \cmark & CLS &  \challengeMark  &  \challengeMark  &  &  & - \\

&  & exFeCAM~\cite{Kushawaha2024Continual} & \cmark  &  &  &  &  & \cmark &  &  & \cmark &  &  & CLS &  \challengeMark  &  &  &  & - \\
\bottomrule
\end{NiceTabular}}

\begin{flushleft}
\smallskip
``Modality'': \cmark \ indicates that the respective modality is included.
\end{flushleft}

\end{table*}

\renewcommand{\aboverulesep}{2pt}

\rev{\mmclMethods ZSCL~\cite{Zheng2023Preventing} calculates the feature similarity between each image (or text) and various texts (or images) in a reference dataset using the fine-tuned and the frozen pre-trained CLIP models.
It then employs KD to match the similarity distributions, thus preserving zero-shot transfer ability (\challengeNum{4})}.
\rev{Similarly, CTP~\cite{Zhu2023CTP} and MulKI~\cite{Zhang2024MultiStage} use distillation to maintain the image-text similarity distribution for cross-modal topology preservation. MulKI~\cite{Zhang2024MultiStage} additionally distills the relation matrices of visual features to maintain intra-modal relationships.}
\citet{Ni2023Continual} demonstrate that continual training of CLIP will cause the Spatial Disorder (SD) issue in vision-language representation and may lead to downgraded performance.
Thus, their proposed method, Mod-X, preserves the spatial distribution of representations between modalities using distillation on the contrastive matrices of the last and current CLIP models.
\rev{MSPT~\cite{Chen2024Continual} captures modality interactions via shared learnable keys in image and text self-attention modules and applies distillation to the attention maps for knowledge maintenance.
To address modality imbalance (\challengeNum{1}), it proposes a gradient modulation strategy to balance the learning of modalities, inspired by OGM-GE~\cite{Peng2022Balanced}.}

\revsec{\summary Figure~\ref{fig:Regularization-based Architectures} provides a summary of typical architectures of modality consistency and relation regularization-based methods.
They effectively preserve knowledge of past models with clear benefits.
A primary advantage is that methods in this category are model-agnostic and easily applicable to various multimodal backbones.
They typically focus on designing sophisticated loss functions rather than altering the model architecture.
Furthermore, modality relation regularization methods that employ relation-based KD offer a natural fit to preserve the crucial cross-modal relationships inherent in paired data, thereby facilitating the development of various MMCL methods in this subcategory and demonstrating great potential for future research.

Despite their advantages, these methods present several limitations.
Firstly, methods usually require full-parameter fine-tuning and demand heavy computation overhead~\cite{Tang2025Mind} (\challengeNum{3}).
Secondly, methods such as ZSCL rely on external reference datasets, which pose practical challenges in real-world scenarios~\cite{Tang2025Mind} and make performance sensitive to the choice of this dataset~\cite{Zhang2024MultiStage}.
Lastly, we observe that the modality consistency regularization methods mainly directly address \challengeNumDesc{0} and rarely tackle other MMCL challenges.
This narrow focus potentially limits the exploration of novel MMCL methods, because their key motivation and techniques are similar to methods in unimodal CL settings.
}

\subsection{Architecture-based Approach}
\label{sec: Continual Learning_Architecture-based}
Architecture-based methods employ an intuitive and direct strategy to learn tasks, by enabling different model parameters to cope with different tasks.
Regularization-based methods share all parameters to learn tasks, making them prone to inter-task interference~\cite{Wang2024Comprehensive}: an issue where remembering old tasks greatly interferes with learning a new task, leading to decreased performance, when the forward knowledge transfer is negative~\cite{Wang2021AFEC}.
In contrast, architecture-based methods reduce inter-task interference by incorporating task-specific components.
\rev{Depending on the factors driving architectural designs, we categorize methods into two types: \textbf{\textit{task-driven architecture}} and \textbf{\textit{modality-driven architecture}}.}
We provide an overview of representative architectures for these two subcategories in \Cref{fig:Architecture-based Architectures}.

\subsubsection{\rev{\textbf{Task-driven Architecture}}}
\label{sec:FA}
\rev{
Task-driven architecture methods design mechanisms in response to new tasks in the MMCL setting.
By dedicating different parameters to individual tasks, these methods inherently reduce inter-task interference and mitigate forgetting.
Typical mechanisms include task-specific parameter allocation within fixed networks and dynamic expansion of new modules.
This architectural flexibility allows some methods to effectively address \textbf{\challengeNumDesc{2}} through specifically designed modules.
Furthermore, by freezing the pre-trained multimodal backbone and only training lightweight new modules, some approaches also address \textbf{\challengeNumDesc{3}}, and \textbf{\challengeNumDesc{4}}.
}
It is worth noting that, if the model has task-specific components and receives the task-ID during testing (the TIL scenario), the primary objective remains to avoid forgetting; however, the model should also effectively learn shared knowledge across tasks and balance performance with computational complexity~\cite{vandeVen2022Three}.

\redsout{\textbf{Representative Unimodal Models.}
HAT~\cite{Serra2018Overcoming} learns near-binary attention vectors for masking, enabling the activation or deactivation of units across different tasks.
Based on the obtained mask, a subset of parameters remains static during training, which helps maintain early knowledge.
}

\rev{\mmclMethods When a new task is introduced, one straightforward strategy is to include task identity as input to control parameter allocation or directly add a new module into the network, i.e., \textit{direct task-driven}. }
A more sophisticated approach is to design a mechanism that adaptively determines how to modify the network for learning new knowledge while maintaining computation efficiency, \rev{i.e., in an \textit{adaptive task-driven} manner. }
Therefore, we group methods based on these two mechanisms in the subsequent paragraphs.

\rev{In \textit{direct task-driven} MMCL methods, tasks and task-specific parameters have a \textit{direct} correspondence.
RATT~\cite{DelChiaro2020RATT} introduces a binary vocabulary mask to selectively inhibit neurons and allocate distinct activations across layers for different tasks, inspired by HAT~\cite{Serra2018Overcoming}.
Alternatively, instead of operating on fixed models, adding modules increases model capacity with each new task, thereby ensuring that performance is not ultimately constrained by the initial capacity~\cite{Peng2021Hierarchical}.
Both MoE-Adapters4CL~\cite{Yu2024Boosting} and DIKI~\cite{Tang2025Mind} freeze the pre-trained CLIP model and introduce task-specific parameter-efficient modules  (\challengeNumTwo{3}{4}). }
MoE-Adapters4CL contains LoRA~\cite{Hu2022LoRA} modules as experts within the MoE framework~\cite{Jacobs1991Adaptive}, along with task-specific routers that determine the weighted aggregation of experts.
\rev{DIKI learns task-specific key and value matrices with residual mechanism.
When the task-ID is unknown, both methods design distribution-based mechanisms to infer the task and select the most relevant modules.
}
CMR-MFN~\cite{Wang2023Confusion} fixes encoders of each modality and adds a modality fusion network for each task in training.
\rev{SS~\cite{Ahrens2023Visually} also uses task-specific modules, such as feed-forward and attention blocks, but proposes different selective specialization strategies to select these modules and analyzes their effectiveness}. %
CLAP~\cite{Jha2024CLAP4CLIP} introduces a visual-guided attention module to align learned text features and pre-trained image features. Moreover, it proposes task-specific adapters to capture task-specific text feature distributions, trained with probabilistic fine-tuning.

Some methods \textit{adaptively decide} when to expand, prune, or alter the network during the training process.
These methods mitigate the increased training costs and redundancy caused by simply adding network parameters for each task.
EProj~\cite{He2023Continual} is proposed alongside TIR~\cite{He2023Continual} (introduced in Section~\texorpdfstring{\protect\hyperlink{mylink TIR}{\getrefnumber{sec:ER}}{}}~), which leverages task similarity scores to determine whether to add a new task-specific module.
If all similarity scores are below a threshold, EProj expands the projection layer in the multimodal base model for the new task, learns task-specific keys, and freezes other modules to prevent forgetting.
VLKD~\cite{Peng2021Hierarchical} constructs a hierarchical recurrent network that expands to learn new knowledge and adaptively deletes less relevant parameters.
MSCGL~\cite{Cai2022Multimodal} is a multimodal graph model with structure-evolving GNN cells, extending the framework of GraphNAS~\cite{Gao2019GraphNAS}. In the search space of aggregation, activation, and correlation operators, MSCGL aims to find the best architecture to learn new tasks.

\subsubsection{\rev{\textbf{Modality-driven Architecture}}}
\label{sec:DA}
\rev{
Modality-driven architecture methods introduce structural changes in response to shifts in modalities, a scenario unique to MMCL.
Unlike other methods that usually assume a fixed set of modalities, this category considers a more realistic situation where the modalities of tasks may vary during the CL process.
By explicitly designing mechanisms for dynamic modality sets, these methods effectively handle \challengeNumDesc{1} and also offer novel ways to manage \challengeNumDesc{2}.
}

\redsout{\textbf{Representative Unimodal Models.}
An early work, namely Progressive Network~\cite{RusuRDSKKPH16}, initializes a new network for each new task. This strategy is explicitly designed to prevent the forgetting of previously learned tasks.
It facilitates knowledge transfer by employing lateral connections to leverage previously acquired features.
}

\rev{\mmclMethods To handle changes in input modalities, SCML~\cite{Song2021Realworld}, as a \textit{modality-driven} method, proposes an architecture as a unified model with a meta-learner and dynamic modality encoders.
}
In this framework, they propose so-called plug networks as dedicated encoders for individual modalities, which map features to the same dimension.
It learns each modality sequentially and utilizes a meta-learner to update the unified model to avoid forgetting. Therefore, it is extensible for accommodating the arrival of new modalities.
The advantage of SCML is that the unified model maps different modalities into a common feature space and avoids explicit alignment between modalities.
ODU~\cite{Sun2021Multimodal} develops classifiers for each task and modality. The data of some modalities may be available in the first task but missing in later tasks. It even trains classifiers of missing modalities, leveraging other modalities as the auxiliary information source.

\revsec{\summary
In \Cref{fig:Architecture-based Architectures}, we illustrate the typical architectures of task-driven and modality-driven architecture-based methods.
To add CL capabilities to multimodal models, designing various architectures is an intuitive strategy and yields various advantages.
Firstly, methods reduce inter-task interference with task-specific parameters and enhance learning capability.
Another important advantage of this category is its flexibility and versatility, with methods designed to address each of the core MMCL challenges in Section~\texorpdfstring{\protect\hyperlink{mylink challenges}{\getrefnumber{sec: Introduction}}{}}~.
Especially when applied to foundation models, an effective strategy is to freeze the pre-trained encoders while only training lightweight modules (\Cref{fig:Architecture-based Architectures}).
Methods adopting this strategy (e.g., MoE-Adapters4CL, DIKI, and EProj) effectively reduce computational costs and preserve zero-shot capabilities  (\mbox{\challengeNumTwo{3}{4}}).

However, extra architectural designs introduce significant drawbacks. Firstly, the model size grows with the arrival of new tasks, leading to increased computation costs and ultimately negating the expected efficiency of continual learning. A typical example is CLAP, whose trainable parameters could potentially exceed those of the pre-trained CLIP~\cite{Jha2024CLAP4CLIP}. Secondly, in CIL scenarios, methods such as MoE-Adapters4CL that rely on task-specific modules often require task-ID discriminators,
Such designs are therefore fragile, as discriminators' ineffectiveness directly leads to poor performance on corresponding tasks~\cite{Lu2024Boosting}.
}

\subsection{Replay-based Approach}
\label{sec: Continual Learning_Replay-based}
Replay-based methods utilize an episodic memory buffer to replay historical instances, such as data samples, from previous tasks, helping to maintain early knowledge while learning new tasks.
This approach of replaying instances avoids the rigid constraints of regularization-based methods and circumvents the complexity of dynamically modifying network architectures in architecture-based methods.
\rev{
    Depending on the strategies to preserve and utilize multimodal information during the replay process, replay-based methods are divided into two sub-directions: \textbf{\textit{natural multimodal replay}} and \textbf{\textit{interactive multimodal replay}}.
}
When learning the $t$-th task, the episodic memory $\mathcal{M}_t$ will be combined with the incoming data $\mathcal{D}_t$. The loss function can be expressed as:
\begin{equation}
    \mathcal{L}_t=\frac{1}{\left|D_{t} \cup \mathcal{M}_{t}\right|} \sum_{(\mathbf{x}_i, y_i) \in\left(D_{t} \cup \mathcal{M}_{t}\right)} \ell(f(\mathbf{x}_i), y_i).
\end{equation}
We depict the representative architectures of these two
subcategories in \Cref{fig:Replay-based Architectures}.

\subsubsection{\rev{\textbf{Natural Multimodal Replay}}}

\label{sec:DR}
\rev{
Natural multimodal replay aims to preserve the natural and inherent relationships between modalities within datasets.
A straightforward approach is to directly replay stored multimodal data pairs or tuples.
For tasks with a looser modality link (e.g., connected only by a shared label space rather than being uniquely paired), some methods generate data for each modality separately.
The natural multimodal relationship is thus preserved implicitly through the same class label.
This generative approach, as an extension of unimodal replay, also avoids the storage requirements and privacy concerns of replaying raw samples.
Methods in this category primarily target \textbf{\challengeNumDesc{0}} through replay of past data information.}

\redsout{\textbf{Representative Unimodal Models.}
Early studies of unimodal direct replay methods have focused on selecting samples based on some heuristic strategies. For instance, Reservoir Sampling~\cite{Vitter1985Random} randomly chooses raw samples. iCaRL~\cite{Rebuffi_2017_CVPR} employs a herding mechanism based on feature representations to ensure class balance.
ER-MIR~\cite{aljundi2019online} selects samples that have a large influence on loss change.
Subsequent work primarily focuses on exploring other selection \mbox{strategies~\cite{Riemer2018Learning,yoon2021online,Choi2024DSLR,Zhou2021Overcoming,Zhang2024Influential}} or optimizing memory \mbox{storage~\cite{aljundi2019online,wang2022memory}}.}

\rev{\mmclMethods }
With multimodal data, an intuitive implementation involves directly selecting and replaying samples from various modalities.
\rev{
For instance, following the sampling strategies from~\cite{Chaudhry2019Tiny} and~\cite{Vitter1985Random}, VQACL~\cite{Zhang2023VQACL}, GMM~\cite{Cao2024Generative} and SAMM~\cite{Sarfraz2024Unimodal} select multimodal samples randomly.}
Experimental results from SAMM~\cite{Sarfraz2024Unimodal} demonstrate that, compared to unimodal replay, multimodal replay significantly enhances the plasticity and stability of the model, thereby achieving a superior stability-plasticity trade-off.

\rev{Once multimodal data is stored in a buffer, some methods naturally integrate with KD, ensuring that the model maintains consistency in various aspects of the old data before and after model updates~\cite{Wang2024Comprehensive}.}
To ensure consistency at the \textit{representation level}, TAM-CL~\cite{Cai2023TaskAttentive} computes the KD loss between the outputs of the last self-attention block from the current student model and the earlier teacher model.
This strategy helps to constrain distribution shifts.
In terms of consistency in \textit{cross-modal interactions}, KDR~\cite{Yang2023Knowledge} utilizes KD to regulate the cross-modal similarity matrix, thereby enhancing the consolidation of cross-modal knowledge.

\rev{When one modality contains a small fixed set of data and is always available, such as templated sentences for classes in CLIP-based image classification, methods focus on replaying the more complex modality.
RAIL~\cite{Xu2024Advancing} employs recursive ridge regression with a memory matrix, theoretically achieving absolute memorization on visual knowledge.}
RAPF~\cite{Huang2024ClassIncremental} generates old class image features, and encourages them to be closer to the text embeddings of old classes and away from those of new classes, thereby mitigating forgetting in CIL.

\subsubsection{\rev{\textbf{Interactive Multimodal Replay}}}
\label{sec:PR}
\rev{
In contrast to natural replay, which preserves inherent multimodal relationships, interactive multimodal replay methods actively learn and utilize modality interaction information when generating data or features for pseudo replay.
These methods typically capture this interaction knowledge through designed modules and prototypes.
This category is particularly effective for addressing \textbf{\challengeNumDesc{2}}.
It may also help alleviate \textbf{\challengeNumDesc{1}}, especially when complex interactions are involved with more than two modalities.
}

\redsout{\textbf{Representative Unimodal Models.}
DGR~\cite{shin2017continual} is a pioneer unimodal work that trains a GAN~\cite{goodfellow2014generative} to generate data samples, which are then replayed during the current model training to retain the previously learned knowledge.
Subsequent research expands this strategy by exploring a variety of generative models~\cite{kemker2017fearnet,riemer2019scalable,ye2020learning} to enhance replay fidelity and scope.
Additionally, some studies shift the focus to the \textit{feature level}~\cite{xiang2019incremental, Hayes2020REMIND, Yang2021Learning, Lao2022TwoStream}, aiming to reinforce feature representations to counteract the issue of forgetting.
}

\rev{\mmclMethods }
With datasets that include various modalities, generating highly correlated data tuples, such as image-question-answer triplets that are both detailed and accurately labeled, usually poses significant challenges.
\rev{To address these difficulties, some studies have focused on learning modality interactions, thus effectively storing multimodal knowledge and generating either substitute or partial data.}
For instance, SGP~\cite{Lei2023Symbolic} maintains scene graphs, which are graphical representations of images, and incorporates a language model for pseudo replay.
IncCLIP~\cite{Yan2022Generative} emphasizes pseudo text replay through the generation of negative texts conditioned on images, which helps better preserve learned knowledge.
\rev{Within the Robot Operating System framework, exFeCAM~\cite{Kushawaha2024Continual} maps tactile and vision data to the same feature space and generates class-wise prototypes for continual object recognition.
ConDU~\cite{Gao2025Enhanced} stores task-specific triggers with whole model updates and class prototypes to recover the model and perform prediction at inference.}
In addition, efforts like FGVIRs~\cite{He2024Continual} and AID~\cite{Cheng2024VisionSensor} specifically tackle issues of modality imbalance.
They employ pseudo-representation and pseudo-prototype replay strategies to enhance classifier discriminability.
They address the inherent challenges in multimodal learning environments where maintaining balance across different types of data is crucial.

\revsec{
\summary
As shown in \Cref{fig:Replay-based Architectures}, within the MMCL setting, replay-based methods offer great flexibility in deciding replay data, as they may opt to replay one or multiple modalities based on the specific design of the model.
Furthermore, the required architectural changes to learn multimodal knowledge and facilitate replay are typically minimal compared to those in architecture-based methods.

However, there are limitations to these methods. Firstly, for many methods in natural multimodal replay, directly utilizing past data induces privacy concerns when the data is sensitive.
Moreover, the reliance on a memory buffer introduces a difficult performance-memory trade-off: a fixed-size buffer often leads to performance degradation over long task sequences, while allowing the multimodal samples in the buffer to grow with tasks results in memory overhead~\cite{wang2023sprompts, Jin2024Calibrating}.
Lastly, some methods generate replay data based on specific data representations (e.g., scene graphs for SGP), which may not be readily available in practice, thus limiting their applicability~\cite{Qian2023Decouple}.
}

\subsection{Prompt-based Approach}
\label{sec: Continual Learning_Prompt-based}

With the rapid development of large models and their application in the CL setting, prompt-based methods have recently emerged to better utilize the rich knowledge acquired during pre-training.
These methods offer the advantage of requiring minimal model adjustments and reducing the need for extensive fine-tuning, unlike previous methods that often require significant fine-tuning or architectural modifications.
The paradigm of prompt-based methods involves modifying the input by applying a few prompt parameters in a continuous space, allowing the model to retain its original knowledge while learning additional task-specific information.
Consequently, they are inherently capable of addressing \textbf{\challengeNumDesc{3}}, and \textbf{\challengeNumDesc{4}} in the MMCL setting.
\rev{Depending on the prompt design strategies, prompt-based methods are categorized into two types: \textbf{\textit{multimodal prompt}} and \textbf{\textit{universal prompt}}.
We present representative architectures of these subcategories in \Cref{fig:Prompt-based Architectures}.}

\revsec{\subsubsection{\textbf{Multimodal Prompt}}
\label{sec:MP}
Multimodal prompt methods are designed to learn prompts for each modality, extending from unimodal methods~\cite{Wang2022Learning, Wang2022DualPrompt}, but they usually go beyond simply learning prompts in parallel. Importantly, methods explore the connections between these prompts during continual training, addressing \textbf{\challengeNumDesc{2}}.
}

\redsout{\textbf{Representative Unimodal Models.}
Early unimodal CL studies primarily concentrate on designing prompt architectures that effectively integrate both general and specific knowledge~\cite{Wang2024Comprehensive}.
L2P~\cite{Wang2022Learning} utilizes a prompt pool shared across all tasks, from which only the most relevant prompts are selected for each input sample during training or inference.
In contrast, DualPrompt~\cite{Wang2022DualPrompt} creates two distinct sets of prompt spaces, accommodating both task-invariant and task-specific prompts.}

\rev{\mmclMethods }
\rev{Existing multimodal prompt-based works operate at various granularities, such as \textit{task-specific} prompts (S-liPrompts~\cite{wang2023sprompts}), and \textit{layer-specific} prompts (CPE-CLIP~\cite{DAlessandro2023Multimodal} and TRIPLET~\cite{Qian2023Decouple}).}
These approaches place greater emphasis on designing prompts that cater to different modalities.
For instance, S-liPrompts introduces a joint language-image prompting scheme that enables the image-end transformer to seamlessly adapt to new domains, while enhancing the language-end transformer's ability to capture more semantic information.
\rev{DPeCLIP~\cite{Lu2024Boosting} also learns separate prompts for the image and language branches, generating them via dedicated attention modules.
Moreover, CPE-CLIP, TRIPLET, MM-Prompt~\cite{Li2025MMPrompt} and HPC~\cite{Jin2024Calibrating} focus more on explicitly considering multimodal interactions during continual prompt learning (\challengeNum{2}).}
CPE-CLIP connects language and vision prompts by defining vision prompts as a function of language prompts; TRIPLET proposes decoupled prompts and prompt interaction strategies to model the complex modality interactions;
\rev{HPC designs calibration mechanisms to enhance consistency alignment of prompts across modalities.}

\revsec{\subsubsection{\textbf{Universal Prompt}}
\label{sec:UP}
Instead of learning prompts for different modalities separately, universal prompt methods aim to learn prompts that encapsulate shared knowledge, either across different modalities or at the task level.
This strategy inherently avoids inconsistency among modality-specific prompts.
Moreover, these methods offer a more computationally efficient option while still achieving competitive performance.}

\rev{\mmclMethods Fwd-Prompt~\cite{Zheng2024AntiForgetting} maintains a single prompt pool shared across all modalities and tasks, and dynamically chooses the most relevant ones based on inputs.
CoLeCLIP~\cite{Li2025Coleclip} utilizes task prompts to capture task-specific features and introduces an attention mask to avoid interferences between prompts and the model's pre-trained features.}

\revsec{\summary We summarize the key architectures of multimodal and universal prompt-based methods in \Cref{fig:Prompt-based Architectures}.}
In the MMCL setting, prompt-based methods may choose to modify the input and learn prompts for the encoders of modalities and/or the modality interaction component.
Depending on the model design, these methods may also facilitate interactions between prompts across different modalities.
\revsec{As a newly emerged paradigm built upon foundation models, the prompt-based approach offers several distinct advantages.
The primary strength is that it successfully preserves the pre-trained knowledge and learn downstream tasks parameter-efficiently (\mbox{\challengeNumTwo{3}{4}}).
Moreover, prompts are flexible to be applied at different granularities, enabling the acquisition of multi-perspective and comprehensive knowledge.

However, this category also presents unique limitations.
Firstly, unlike hard prompts with human-readable words, soft learnable prompts lack explainability, making it difficult to understand how they evoke pre-trained knowledge.
Secondly, their effectiveness relies heavily on the underlying pre-trained models, unlike methods in other categories, which also work with models trained from scratch.
Thirdly, prompts with restricted length limit their learning and expressive capabilities~\cite{Gao2025Enhanced}.
Lastly, for methods like S-liPrompts and CoLeCLIP, the prompt parameters scale linearly with the number of tasks, raising concerns in computational and memory costs for long task sequences (\challengeNum{3} reintroduced)~\cite{Li2025Coleclip}.
}
\renewcommand{\aboverulesep}{0pt}

\begin{table*}[t]
\centering
\caption{\rev{A summary of MMCL benchmarks.}}

\resizebox{\linewidth}{!}{
        \begingroup  %
        \renewcommand{\arraystretch}{0.95}  %
\begin{NiceTabular}{l|ccccc|ccccc|c|c}
\CodeBefore
\rowcolors{3}{blue!10}{}[cols=1-13,restart]
\Body
\toprule
\multirow{2.5}{*}{\textbf{Name}} & \multicolumn{5}{c|}{\textbf{MMCL Scenario}} & \multicolumn{5}{c|}{\textbf{Modality}} & \multirow{2.5}{*}{\textbf{Task}} &  \multirow{2.5}{*}{\textbf{Code}}\\
\cmidrule{2-11} 
 & CIL & DIL & XDIL & TIL & MDTIL &   Vision & Language & Audio & Acceleration & Gyroscope  &   \\ \midrule
\climb \cite{Srinivasan2022CLiMB} &  &  &  & \cmark & \cmark & \cmark & \cmark &  &  &  &   CLS &    \href{https://github.com/GLAMOR-USC/CLiMB}{Link}  \\
CLOVE \cite{Lei2023Symbolic} &  &  & \cmark &  &  & \cmark & \cmark &  &  &  &   GEN &    \href{https://github.com/showlab/CLVQA}{Link}  \\
IMNER, IMRE \cite{Chen2024Continual} & \cmark &  &  &  &  & \cmark & \cmark & &  &  &    CLS &  \href{https://github.com/zjunlp/ContinueMKGC}{Link}  \\
MTIL \cite{Zheng2023Preventing} &  &  &  & \cmark &  & \cmark & \cmark &  &  & &   CLS &  \href{https://github.com/Thunderbeee/ZSCL}{Link}  \\
VLCP \cite{Zhu2023CTP} & \cmark &  &  &  &  & \cmark & \cmark &  &  &  & RET &   \href{https://github.com/KevinLight831/CTP}{Link}  \\
LILAC \cite{Ahrens2023Visually} &  &  &  & \cmark &  & \cmark & \cmark &  &  &  & RET &   \href{https://github.com/ky-ah/selective-lilac}{Link}  \\
MMCL \cite{Sarfraz2024Unimodal} & \cmark & \cmark &  &  &    & \cmark &  & \cmark &  & &   CLS &  \href{https://github.com/NeurAI-Lab/MultiModal-CL}{Link}  \\
CEAR \cite{Xu2024Continual} & \cmark &  &  &  &  & \cmark &  &    & \cmark & \cmark & CLS &  \href{https://github.com/Xu-Linfeng/UESTC_MMEA_CL_main}{Link}  \\

\bottomrule
\end{NiceTabular}

        \endgroup  %
}

\label{tab:summarization benchmarks}
\end{table*}

\renewcommand{\aboverulesep}{2pt}

\section{Datasets and Benchmarks}
\label{sec:Benchmark}
In this section, we provide an overview of current datasets and benchmarks in MMCL.
\rev{
    There are three primary construction approaches: 
    (\hyperref[sec:Benchmark Original]{A}) collecting new datasets that are dedicated to MMCL;
    (\hyperref[sec: Benchmarking on Several Datasets]{B}) linking multiple well-known datasets that are initially designed for non-CL tasks;
    and (\hyperref[sec: Benchmarking on a Partitioned Dataset]{C}) partitioning a single dataset into multiple subsets to simulate tasks in the MMCL setting~\cite{Xu2024Continual}. 
}

\Cref{tab:summarization benchmarks} summarizes MMCL benchmarks covering various CL scenarios, modalities, and task types. 
We introduce them as follows if codes are publicly accessible.

\subsection{Benchmarking on an Original Dataset}
\label{sec:Benchmark Original}
\rev{
    In this subsection, we summarize three benchmarks with newly collected datasets dedicated to the MMCL setting. Compared to the other two approaches, which manually alter existing datasets, these original datasets help evaluate and scale for more practical MMCL settings.
}
\citet{Zhu2023CTP} utilize E-commerce data to construct the first vision-language continual pre-training dataset P9D and establish the VLCP benchmark for cross-modal retrieval and multimodal retrieval. 
P9D contains more than one million image-text pairs of real products and is partitioned into 9 tasks by industrial categories. 
\rev{\citet{Ahrens2023Visually} introduce the LILAC benchmark with two datasets: LILAC-2D generated based on minigrid~\cite{minigrid} and LILAC-3D based on Ravens~\cite{zeng2021transporter} and \textsc{CLIPort}~\cite{shridhar2022cliport}, for language-instructed tasks grounded in simulated visual environments. Baseline CL methods (ER~\cite{Chaudhry2019Tiny}, EWC~\cite{Kirkpatrick2017Overcoming}) under LILAC-2D yield suboptimal performance, with an average accuracy more than 20\% lower than the newly proposed MMCL method SS~\cite{Ahrens2023Visually}. }
\citet{Xu2024Continual} collect video and sensor data from ten participants wearing smart glasses. They construct the dataset UESTC-MMEA-CL, the first multimodal dataset for continual egocentric activity recognition, with modalities of vision, acceleration, and gyroscope. They also establish a benchmark, CEAR, with three baseline CL methods, namely EWC~\cite{Kirkpatrick2017Overcoming}, LwF~\cite{Li2017Learning} and iCaRL~\cite{Rebuffi_2017_CVPR}. 
Results demonstrate that replay-based iCaRL is more effective in alleviating forgetting than replay-free methods EWC and LwF. 
Nonetheless, exploring replay-free strategies remains promising and important, as replay-based methods are not always applicable due to considerations such as privacy concerns~\cite{Xu2024Continual}. 
\citet{Xu2024Continual} use TBW~\cite{Kazakos2019EPICFusion}-like midfusion to fuse multimodal features, achieving better results than using single modality data in the non-CL setting. 
However, in the MMCL setting, the performance with multimodal data (vision and acceleration) is inferior to that with unimodal data (vision), even with CL methods incorporated. 
These results highlight the necessity for further research in MMCL methods to improve the fusion of modality information while preventing forgetting.

\subsection{Benchmarking on Several Datasets}
\label{sec: Benchmarking on Several Datasets}
\rev{
    We outline three benchmarks that utilize various existing datasets as tasks in the MMCL framework. These independently collected datasets exhibit drastically different distributions and are challenging for model learning.
}
\climb\ \cite{Srinivasan2022CLiMB} benchmarks with four vision-language tasks (VQAv2~\cite{Goyal2017Making}, NLVR2~\cite{Suhr2019Corpus}, SNLI-VE~\cite{Xie2019Visual}, and VCR~\cite{Zellers2019Recognition}), five language-only tasks (IMDb~\cite{Maas2011Learning}, SST-2~\cite{Socher2013Recursive}, HellaSwag~\cite{Zellers2019HellaSwag}, CommonsenseQA~\cite{Talmor2019CommonsenseQA}, and PIQA~\cite{Bisk2020PIQA}) and four vision-only tasks (ImageNet-1000~\cite{Russakovsky2015ImageNet}, iNaturalist2019~\cite{VanHorn2018INaturalist}, Places365~\cite{Mahajan2018Exploring}, and MS-COCO object detection~\cite{Lin2014Microsoft}).
\climb\ treats each task as a classification task and consists of two phases within the CL process.
In upstream continual learning, the model is trained on vision-language tasks with various candidate CL algorithms.
In downstream low-shot transfer, after training on the $i$-th upstream task and saving checkpoints, for each task of the training data of the remaining upstream tasks and unimodal tasks, the model is fine-tuned on the checkpoints with a fraction of the task data.
The \climb\ benchmark results demonstrate that common CL algorithms (ER~\cite{Chaudhry2019Tiny}, EWC~\cite{Kirkpatrick2017Overcoming}) are able to alleviate forgetting. However, they may hurt downstream task learning, compared to direct fine-tuning. These results underscore the need for further research on MMCL methods.
CLOVE~\cite{Lei2023Symbolic} splits data from GQA~\cite{Hudson2019GQA} 
into six subsets representing different scenes, such as \textit{workplaces}
for the CLOVE-scene CL setting, following the taxonomy in SUN397~\cite{Xiao2010SUN}. 
Additionally, CLOVE collects six functions, such as \textit{object recognition}, 
for the CLOVE-function CL setting, using data from GQA~\cite{Hudson2019GQA}, CRIC~\cite{Gao2023CRIC}, and TextVQA~\cite{Singh2019VQA}.
CLOVE evaluates the performance of different methods on continual learning of different VQA tasks. 
Lastly, MTIL~\cite{Zheng2023Preventing} is a challenging benchmark consisting of eleven image classification tasks from different domains, including Aircraft~\cite{Maji2013FineGrained}, 
Caltech101~\cite{Fei-Fei2004Learning}, 
CIFAR100~\cite{Krizhevsky2009Learning}, 
DTD~\cite{Cimpoi2014Describing}, 
EuroSAT~\cite{Helber2019EuroSAT}, 
Flowers~\cite{Nilsback2008Automated}, 
Food~\cite{Bossard2014Food101}, 
MNIST~\cite{Deng2012MNIST}, 
OxfordPet~\cite{Parkhi2012Cats}, 
StanfordCars~\cite{Krause20133D}, and 
SUN397~\cite{Xiao2010SUN}.

\subsection{Benchmarking on a Partitioned Dataset}
\label{sec: Benchmarking on a Partitioned Dataset}
\rev{
    In MMCL benchmarking, the most straightforward approach is to utilize a well-known dataset initially designed for non-CL settings and partition it into multiple subsets to simulate a sequence of tasks. 
    Such simplicity facilitates rapid initial progress in MMCL research, with three benchmarks developed in this manner.
}
The IMNER benchmark~\cite{Chen2024Continual} utilizes the Twitter-2017 MNER dataset (constructed by~\cite{Lu2018Visual} and preprocessed by~\cite{Yu2020Improving}) and splits it by categories to simulate the CIL scenario. The IMRE benchmark~\cite{Chen2024Continual} partitions the MEGA MRE dataset~\cite{Zheng2021Multimodal} into 10 subsets for the CIL scenario. 
MMCL~\cite{Sarfraz2024Unimodal} is a benchmark that contains audio and visual modalities for classification. It partitions the VGGSound dataset~\cite{Chen2020Vggsound} to simulate CIL and DIL scenarios.

\revsec{
\section{Discussion}
\label{sec: Discussion}
In this section, we present an in-depth discussion of research trends, foundation models in MMCL, and the evolved stability-plasticity trade-off. 
\subsection{Observation of Research Trends}
\label{sec: Observation of Research Trend}

As MMCL methods continue to emerge, we observe and summarize two pivotal research trends. 
The first trend is a significant shift in the challenges that MMCL methods focus on, which has progressively distinguished MMCL from traditional CL and fostered the development of this field.
The second trend is a transition from training models from scratch to adapting pre-trained foundation models, providing stronger CL capabilities and better practical utility in real-world scenarios.

\subsubsection{Challenges}
Inherited from traditional unimodal CL, whose major challenge is catastrophic forgetting, early MMCL methods naturally only focus on \challengeNumDesc{0} (e.g., \cite{DelChiaro2020RATT}). 
Subsequently, MMCL works start to identify and address new challenges in this field (Section~\texorpdfstring{\protect\hyperlink{mylink challenges}{\getrefnumber{sec: Introduction}}{}}\ ).
This evolution has advanced MMCL as a unique research area, differentiating it from traditional CL.
Consequently, many recent works address these new challenges, yielding improved performance and alleviating the more severe forgetting that occurs in the multimodal setting~\cite{Xu2024Continual, Sarfraz2024Unimodal}.

\subsubsection{Backbones}
Mirroring a broader trend in the ML community, backbones in MMCL have evolved from small, randomly initialized models to large foundation models.
Many early approaches design custom architectures and train them from scratch (e.g., \cite{DelChiaro2020RATT, Sun2021Multimodal, Peng2021Hierarchical}). 
The rise of foundation models has profoundly changed the landscape of the field, with a majority of recent MMCL methods leveraging them as powerful backbones (\Cref{tab:summarizations}). 
This adoption equips models with a head start in pre-trained knowledge, greater learning capacity, and better performance. We discuss further details in the next section.

\subsection{Foundation Models in MMCL}
\label{sec: Foundation Models in MMCL}
In this subsection, we explore the pivotal role of foundation models and discuss MMCL methods based on them.

\subsubsection{Foundation Models}
Pre-trained on large-scale multimodal datasets, foundation models such as CLIP~\cite{Radford2021Learning}, BLIP2~\cite{li2023blip2}, and GLIP~\cite{Li2022Grounded} have acquired extensive general knowledge. 
Performing CL on top of these MM backbones thus provides a significant head start. 
Simultaneously, these foundation models cannot remain static but require CL to adapt to the dynamically evolving world. 
Full retraining with both pre-trained and new data is computationally prohibitive, underscoring the necessity and importance of MMCL. 
This establishes a mutually beneficial relationship between foundation models and MMCL methods.

Importantly, foundation models offer remarkable flexibility as backbones, as they are inherently multimodal. 
Besides those pre-trained on multimodal data, even unimodal-trained models, such as LLMs, can be transformed into powerful multimodal learners (MLLMs). By extracting and feeding features from other modalities, LLMs are able to adapt to multimodal tasks. Consequently, some MMCL methods do not directly use off-the-shelf MM backbones but instead assemble custom ones (e.g., MM-Prompt~\cite{Li2025MMPrompt} combines LXMERT~\cite{Tan2019LXMERT} and T5~\cite{Raffel2020Exploring}). 
The selection and combination of foundation models in MMCL are diverse, fostering a flourishing of various MMCL techniques built upon them, which we detail next.

\subsubsection{MMCL Methods}
The rise of foundation models has enriched the landscape of MMCL, with a primary focus on vision-language modalities (\Cref{tab:summarizations}). %
At the \textit{task level}, the choice of backbone often links to the task type.
Many CLIP-based MMCL methods focus on image classification, which involves matching images with templated class sentences from multiple domains (e.g., \cite{Zheng2023Preventing, Yu2024Boosting, Xu2024Advancing}). 
Meanwhile, MLLM-based approaches are frequently applied to generative tasks like continual VQA (e.g., \cite{Zhang2023VQACL, Lei2023Symbolic, Li2025MMPrompt}). 
At the \textit{technique level}, methods focus on preserving and utilizing knowledge of foundation models when learning new knowledge.
They protect existing knowledge at the parameter level (regularization-based) and the data level (replay-based).
Meanwhile, some techniques aim to elicit knowledge from foundation models through learned prompts (prompt-based) and facilitate new knowledge acquisition through architectural modifications (architecture-based).
Prompt-based category, in particular, is a newly emerged paradigm whose design is uniquely suited to powerful foundations. Concurrently, other established categories also further advance with new methods designed for these foundations.

\subsubsection{Evaluations}
While the diversity of foundation models facilitates the development of MMCL methods, as discussed in the above two sections, it highlights an evaluation issue.
For a fair comparison, methods should be adapted to and evaluated on the same backbone. 
We note that for some datasets, the performance ranking of methods can invert with backbone changes 
(e.g., EProj~\cite{He2023Continual} versus Fwd-Prompt~\cite{Zheng2024AntiForgetting} on BLIP2~\cite{li2023blip2} and InstructBLIP~\cite{dai2023instructblip}).
This phenomenon shows the backbone dependency of these foundation model-based methods and highlights the need to enhance robustness.

\subsection{Stability-plasticity Trade-off}
\label{sec: Stability-plasticity Trade-off}
In this subsection, we present an in-depth discussion of the stability-plasticity trade-off dilemma, with an emphasis on its evolution under MMCL methods with foundation models.
\subsubsection{Stability}
Stability refers to a model's ability to retain previously learned information~\cite{Mermillod2013Stabilityplasticity}. 
For MMCL methods with small models trained from scratch, stability primarily concerns knowledge acquired from the task sequence, as randomly initialized models begin with no prior knowledge. 
However, the incorporation of foundation models expands the scope of stability to include a crucial second dimension: preserving their pre-trained general knowledge (\challengeNum{4}).  %
Forgetting pre-trained knowledge in training leads to loss of \textbf{zero-shot capabilities} and degrades performance on future tasks, i.e., negative forward transfer~\cite{Zheng2023Preventing, Zheng2024AntiForgetting}. 
On the other hand, while simply using the original pre-trained checkpoint ensures perfect stability, it is inadequate because direct fine-tuning and MMCL methods allow new knowledge acquisition and achieve better performance on the current task.  %

Notably, foundation models possess a high degree of inherent stability and demonstrate mitigated forgetting. Even after direct fine-tuning on new tasks, despite performance degradation, they often maintain the performance of old tasks at a reasonable level~\cite{Zheng2023Preventing}, rather than regressing to near-random guessing like small models~\cite {Kirkpatrick2017Overcoming}. Their vast general knowledge and large number of parameters prevent a complete loss of knowledge. 
Therefore, building MMCL methods on foundation models provides a more robust starting point, which aligns with the research trend discussed in Section~\ref{sec: Observation of Research Trend}.

\subsubsection{Plasticity}
Plasticity is defined as the capacity to acquire new knowledge~\cite{Mermillod2013Stabilityplasticity}.
While small models are computationally inexpensive, they face large limitations due to their model capacity, especially with long task sequences. 
Large foundational models are better at adapting to incoming tasks with pre-trained knowledge and a large number of parameters. 
However, full-parameter fine-tuning comes at a high training cost, posing practical concerns (\challengeNum{3}). 
Consequently, some MMCL methods focus on controlled adaptation with a manageable number of trainable parameters.
By freezing most parameters and training only a small subset, these methods achieve plasticity and address \challengeNum{3}.

\subsubsection{Trade-off}
Plasticity enables the acquisition of new information, but it also causes disruption, leading to forgetting. Conversely, excessive stability, while preventing forgetting of previous tasks, is undesirable if it impedes adaptation~\cite{Mermillod2013Stabilityplasticity}. 
A trade-off between the two is therefore necessary.
For small models with unimodal data, methods such as EWC~\cite{Kirkpatrick2017Overcoming} and Progressive Network~\cite{RusuRDSKKPH16} address this issue by regularizing on parameters or initializing new networks for each task. 
However, multimodal data inevitably increases the parameters required for learning.
For large foundation models, these traditional strategies become challenging to scale due to computation and storage burdens. Therefore, more efficient MMCL solutions are needed to achieve this balance.

With foundation models, the situation is further complicated by the need to manage an additional trade-off between plasticity and the stability of pre-trained knowledge, which gives powerful zero-shot generalization capabilities~\cite{Zhang2024MultiStage, Tang2025Mind}.
Some methods achieve this balance by freezing most parameters or keeping a copy of the original model, while using a small set of trainable parameters to learn new knowledge. 
The key idea in modern methods for managing this new trade-off is to decouple general and task-specific knowledge. 
Architecture and prompt-based methods achieve this explicitly by design, storing new knowledge in dedicated modules or prompts. 
In contrast, regularization and replay-based methods, without additional strategies, typically require full-parameter fine-tuning and entangle these two knowledge types within a shared parameter space, leading to inter-task interference.

Beyond the overall multimodal trade-off, MMCL should also consider the trade-off for each modality. 
Due to modality imbalance (\challengeNum{1}), modalities do not learn or forget in a synchronized manner~\cite{He2024Continual}. 
For instance, language features naturally dominate vision features in continual VQA~\cite{Li2025MMPrompt}. 
This crucial aspect is currently addressed by only a few MMCL methods (e.g., \cite{Chen2024Continual, Li2025MMPrompt}). 

In summary, while various methods have been proposed to manage the evolved stability-plasticity trade-off, MMCL is still in its early stages. This dilemma remains a critical issue that requires further research and development.

}

\section{Future Directions}
\label{sec:Future Direction}
\newcommand{\nodeFutureFstLevel}{1.4cm}
\newcommand{\nodeFutureSndLevel}{6.1cm}

\begin{figure}[t]

  \centering
  
  \tikzset{
          my node/.style={
              draw,
              align=center,
              thin,
              text width=1.2cm, 
              rounded corners=3,
          },
          my leaf/.style={
              draw,
              align=left,
              thin,
              text width=8.5cm, 
              rounded corners=3,
          }
  }
  \forestset{
    every leaf node/.style={
      if n children=0{#1}{}
    },
    every tree node/.style={
      if n children=0{minimum width=1em}{#1}
    },
  }
  \begin{forest}
      nonleaf/.style={font=\bfseries\scriptsize},
       for tree={%
          every leaf node={my leaf, font=\scriptsize},
          every tree node={my node, font=\scriptsize, l sep-=4.5pt, l-=1.pt},
          anchor=west,
          inner sep=2pt,
          l = 8pt,
          l sep=8pt, %
          s sep=4pt, %
          fit=tight,
          grow'=east,
          edge={ultra thin},
          parent anchor=east,
          child anchor=west,
          if n children=0{}{nonleaf}, 
          edge path={
              \noexpand\path [draw, \forestoption{edge}] (!u.parent anchor) -- +(4pt,0) |- (.child anchor)\forestoption{edge label};
          }, %
          if={isodd(n_children())}{
              for children={
                  if={equal(n,(n_children("!u")+1)/2)}{calign with current}{}
              }
          }{}
      }
      [Future Directions, draw=gray, fill=gray!15, anchor=center,  parent anchor=south, rotate=90,  text width=2cm, text=black
      [Enhanced MMCL Datasets and Benchmark Evaluations \\ ({\cref{sec: Improved MMCL Datasets and Benchmark Evaluations}}), color=brightlavender, fill=brightlavender!15, text width=\nodeFutureFstLevel, text=black
            [{
                Improved Modality Quantity \& Quality
(\cref{sec: Improved Modality Quantity Quality})
                }, color=brightlavender, fill=brightlavender!15, text width=\nodeFutureSndLevel, text=black],
            [{
                Establishment of Large-scale MMCL Datasets and Benchmarks
(\cref{sec: Establishment of Large-scale MMCL Datasets and Benchmarks})
                }, color=brightlavender, fill=brightlavender!15, text width=\nodeFutureSndLevel, text=black],
            [{
                Proposition and Evaluation of MMCL Metrics
(\cref{sec: Proposition and Evaluation of MMCL Metrics})
                }, color=brightlavender, fill=brightlavender!15, text width=\nodeFutureSndLevel, text=black],
          ]
      [Advanced MMCL Techniques \\ (\cref{sec: Advanced MMCL Techniques}), color=lightgreen, fill=lightgreen!15, text width=\nodeFutureFstLevel, text=black
            [{
                Generalization to Missing Modalities
(\cref{sec: Generalization to Missing Modalities})
                }, color=lightgreen, fill=lightgreen!15, text width=\nodeFutureSndLevel, text=black],
            [{
                Novel Modality Interaction Strategies
(\cref{sec: Better Modality Interaction Strategies})
                }, color=lightgreen, fill=lightgreen!15, text width=\nodeFutureSndLevel, text=black],
            [{
                Parameter-efficient Fine-tuning MMCL Methods
(\cref{sec: Parameter-efficient Fine-tuning MMCL Methods})
                }, color=lightgreen, fill=lightgreen!15, text width=\nodeFutureSndLevel, text=black],
            [{
                Better Pre-trained MM Knowledge Maintenance
(\cref{sec: Better Pre-trained MM Knowledge Maintenance})
                }, color=lightgreen, fill=lightgreen!15, text width=\nodeFutureSndLevel, text=black],
            [{
                Development of Prompt-based MMCL Methods
(\cref{sec: Prompt-based MMCL Methods})
                }, color=lightgreen, fill=lightgreen!15, text width=\nodeFutureSndLevel, text=black],
            [{
                Trustworthy Multimodal Continual Learning
(\cref{sec: Trustworthy Multimodal Continual Learning})
                }, color=lightgreen, fill=lightgreen!15, text width=\nodeFutureSndLevel, text=black],
        ]
      ]
  \end{forest}
  \caption{\rev{Taxonomy of future directions of MMCL.}}
  \label{fig: Future Direction}
  \end{figure}

With the rapid advancement of multimodal models, MMCL has become an active and promising research topic.
\rev{As illustrated in \Cref{fig: Future Direction}, we outline future research directions along two primary areas:
(\hyperref[sec: Improved MMCL Datasets and Benchmark Evaluations]{A}) the development of more comprehensive datasets and benchmark evaluations, and
(\hyperref[sec: Advanced MMCL Techniques]{B}) the design of novel and effective MMCL techniques.
}

\subsection{Enhanced MMCL Datasets and Benchmark Evaluations}
\label{sec: Improved MMCL Datasets and Benchmark Evaluations}

\subsubsection{Improved Modality Quantity \& Quality}
\label{sec: Improved Modality Quantity Quality}
Our summarization in \Cref{tab:summarization other modalities} reveals that only a few MMCL methods focus on modalities other than vision and language. Therefore, there is huge space for further research on incorporating more modalities. Similarly, developing benchmarks for more modalities is important for this field.
Moreover, modalities are not limited to those listed in \Cref{tab:summarization other modalities} and may include biosensors~\cite{Cui2020Advancing}, genetics~\cite{Libbrecht2015Machine}, and others~\cite{Acosta2022Multimodal}, thereby enhancing support for emerging challenges, in fields such as AI for science research. 
Various multimodal methods and applications ~\cite{Chen2021Multimodal, Xiong2024MoME, Liu2020MultiLevel, Liu2022Attentionlike} can be extended into the continual learning framework to enhance their learning capabilities.
With the introduction of more modalities, it will be increasingly imperative to address \textit{data-level modality imbalance}, i.e., \challengeNum{1}, which, as shown in \Cref{tab:summarizations} and \Cref{tab:summarization other modalities}, has been addressed by only a few MMCL methods.
Furthermore, due to the discrepancy among distributions and quality of different modalities, the modality with better performance may dominate optimization, leaving other modalities under-optimized~\cite{Peng2022Balanced}.
Hence, addressing \textit{parameter-level modality imbalance} is also crucial. 
Developing specific strategies to balance modalities helps mitigate the forgetting issue~\cite{He2024Continual}, making it a promising research direction.

\revsec{\subsubsection{Establishment of Large-scale MMCL Datasets and Benchmarks}
\label{sec: Establishment of Large-scale MMCL Datasets and Benchmarks}
Many current MMCL datasets are constructed by simply linking or adapting existing academic datasets, as mentioned in Section~\ref{sec:Benchmark}.
While these datasets are relatively straightforward to set up and facilitate rapid initial progress for MMCL research, their limitations in scalability and representativeness become apparent as the field develops.
These datasets often lack the scale and real-world complexity necessary for studying practical MMCL settings.
Specifically, real-world data is long-tail distributed while manually constructed datasets typically have balanced classes or the same sample size for different tasks~\cite{Zhu2023CTP}.
Moreover, improper data partitioning strategies due to a lack of information may not effectively simulate the continual environment~\cite{Zhu2023CTP}.
Consequently, the need for new large-scale datasets for MMCL is increasingly urgent.
Although such dedicated datasets are currently scarce, they already provide important insights through benchmark results, as summarized in Section~\ref{sec:Benchmark Original}.
Therefore, we advocate for the establishment of more large-scale MMCL datasets and benchmarks as an important future direction, which will significantly advance MMCL research.}

\revsec{\subsubsection{Proposition and Evaluation of MMCL Metrics}
\label{sec: Proposition and Evaluation of MMCL Metrics}
The new MMCL challenges require dedicated, fine-grained metrics for a deeper analysis of model capabilities and limitations.
For initial modality-oriented metrics that have been proposed (Section~\texorpdfstring{\protect\hyperlink{mylink MMCL metrics}{\getrefnumber{sec: Evaluation Metric}}{}}~), we advocate for their wider adoption and systematic evaluation across more methods.
Moreover, we call for the development of additional metrics to analyze the degree of challenge resolution from multiple perspectives.
The development and evaluation of MMCL metrics will further advance this field from its early stage to a more mature phase.}

\subsection{Advanced MMCL Techniques}
\label{sec: Advanced MMCL Techniques}

\revsec{\subsubsection{Generalization to Missing Modalities}
\label{sec: Generalization to Missing Modalities}
A key real-world challenge is handling missing modalities (\challengeNum{1}), which introduces practical difficulties to models' robustness. 
It may occur at the sample level for individual data points, or more severely, at the task level, where an entire task lacks a modality.
Models require systematic designs to address this issue, without assuming the completeness of all modalities. 
Developing effective strategies to capture the intrinsic connection between tasks and modalities helps improve model performance~\cite{Sun2021Multimodal}.
A promising starting point for future research is to extend non-CL methods with compositional generalization capabilities, such as generating missing modality features~\cite{Wang2023MultiModal, Raj2024Optimizing}, to the CL scenarios.}

\subsubsection{Novel Modality Interaction Strategies}
\label{sec: Better Modality Interaction Strategies}
As we have just mentioned, there are only a few MMCL methods that incorporate more than two modalities.
Modality interaction, especially modality alignment, may be more complicated with three or more modalities, i.e., \challengeNum{2}.
Furthermore, many existing MMCL methods simply fuse modalities within neural architectures without a deeper understanding or analysis of their mutual influence on learning. 
Thus, it will be interesting and promising to measure such inter-modality influence~\cite{xue2021probing,Xu2023Multimodal} for more fine-grained multimodal interaction.

\subsubsection{Parameter-efficient Fine-tuning MMCL Methods}
\label{sec: Parameter-efficient Fine-tuning MMCL Methods}
Parameter-efficient fine-tuning (PEFT) methods offer an effective solution to optimize training costs, i.e., addressing \challengeNum{3}, by reducing the number of trainable parameters while achieving comparable or better performance than full-parameter fine-tuning to the large models~\cite{Hu2022LoRA,Ding2022Delta}. 
While prompt-based methods are parameter-efficient, in \Cref{tab:summarizations}, we observe that in other categories, only a few methods utilize PEFT techniques. 
CLAP~\cite{Jha2024CLAP4CLIP} also mentions this as its future work.
Therefore, given numerous PEFT methods emerging in recent years~\cite{Lialin2023Scaling, Zhong2024PanDa}, employing them to reduce training costs for MMCL methods is a worthy direction.
Furthermore, beyond the straightforward application of existing PEFT methods, a promising direction is to propose new PEFT methods specifically for the MMCL setting, and to seamlessly integrate them with other MMCL techniques.

\subsubsection{Better Pre-trained MM Knowledge Maintenance}
\label{sec: Better Pre-trained MM Knowledge Maintenance}
As many MMCL methods are armed with powerful MM backbones, it is naturally desirable to memorize their pre-trained knowledge during training. 
Forgetting pre-trained knowledge may significantly hurt future task performance~\cite{Zheng2024AntiForgetting,Zheng2023Preventing}. 
We observe that few methods in \Cref{tab:summarizations}, aside from prompt-based ones, explicitly prioritize maintaining pre-trained knowledge, i.e., addressing \challengeNum{4}, as one of their key goals. 
Moreover, this is particularly challenging for replay-based methods that usually rely on quick adaptation to old data samples for knowledge retention. However, for certain pre-trained models like CLIP, the pre-trained data is private~\cite{Zheng2023Preventing}, which makes the target difficult yet promising for future research.

\subsubsection{Development of Prompt-based MMCL Methods} 
\label{sec: Prompt-based MMCL Methods} 
As discussed in Section~\ref{sec: Continual Learning_Prompt-based}, prompt-based MMCL methods effectively address \challengeNumDesc{3}, and \challengeNumDesc{4}.
However, as shown in \Cref{tab:summarizations}, we note that prompt-based MMCL methods are currently the least represented category. 
Recently, prompt learning techniques are gaining traction in the non-CL setting for multimodal models~\cite{Zhou2022Learning, Khattak2023Maple}. Moreover, there are popular prompt tuning methods that combine learning with high-quality templates~\cite{Yao2024CPT}.
Extending these methods to the MMCL setting facilitates the efficient and effective utilization of pre-trained models. 
Given that the prompt-based category is still in its infancy, there is significant potential for further research and development.

\subsubsection{Trustworthy Multimodal Continual Learning}
\label{sec: Trustworthy Multimodal Continual Learning}
As public concern for model safety and privacy increases, and governments impose more related regulations, the demand for trustworthy models is escalating.  
To ensure safety, models should prevent and detect \textit{hallucination} and \textit{misinformation} to mitigate risks when deployed in the real world~\cite{Yang2024Give, Apostol2024ContCommRTD}.
Although there have been related studies in LLMs~\cite{Liu2024Preventing}, such issues in MMCL have not been explored yet.
A study on multimodal LLMs demonstrates that fine-tuning these models on a single dataset already causes hallucination~\cite{Zhai2023Investigating}.
While the model learns knowledge during the CL process, it should also continue to control and mitigate hallucination and misinformation.
Thus, an important and interesting future direction is to extend such studies in MMCL.
Moreover, models require \textit{safety alignment} with human expectations and social standards to avoid misleading and harmful content~\cite{Liu2024Direct}. However, the concept of safety changes and evolves over time. Thus, incorporating MMCL techniques into multimodal models is essential to address this changing concept.
Regarding privacy, techniques such as \textit{federated learning} (FL) enable the server model to learn knowledge of all clients' data without sharing their raw data. FL techniques also help enhance model robustness, keeping it stable even under extreme conditions such as malicious attacks intended to compromise the model~\cite{Zhang2023Survey, Lyu2024Privacy}.
With numerous federated continual learning (FCL) methods, it would be a promising direction to extend FCL methods to the MMCL setting, so that models maintain knowledge of previous tasks and fuse new knowledge from multimodal data of clients with privacy protected~\cite{Yang2024Federated}.

\section{Conclusion}
\label{sec:Conclusion}
In this work, we present an up-to-date multimodal continual learning survey. 
We provide a structured taxonomy of MMCL methods, essential background knowledge, a summary of datasets and benchmarks, and discuss two novel MMCL scenarios for further study. 
We categorize existing MMCL works into four categories, i.e., regularization-based, architecture-based, replay-based, and prompt-based methods, with detailed subcategories described. 
We also provide representative architecture illustrations for all categories. Our detailed review highlights the key features and innovations of these MMCL methods.
Additionally, we outline promising future research directions in this rapidly evolving field, offering discussions on potential areas for further investigation and exploration.
We anticipate that the development of MMCL will further enhance models to exhibit more human-like capabilities. This enhancement includes the ability to process multiple modalities at the input level and acquire diverse skills at the task level, thereby bringing us closer to realizing general-purpose intelligence in this multimodal and dynamic world. 

\section*{Acknowledgment}

This work was supported in part by the Research Grants Council of the Hong Kong Special Administrative Region, China (CUHK 2410072, RGC R1015-23; CUHK 2300246, RGC C1043-24G), and in part by NSF under grants III-2106758, and POSE-2346158.
The icons are from PowerPoint and \href{https://www.flaticon.com/}{Flaticon}. The robots in \Cref{fig: non-tech Illustration of CL and MMCL} are generated by \href{https://www.bing.com/images/create}{AI Image Creator in Bing}.
We sincerely thank Professor Tat-Seng Chua at the National University of Singapore for his invaluable and insightful suggestions regarding this paper.
We thank the anonymous reviewers for their constructive comments.

\footnotesize{
\bibliography{MyLibrary,ref_short}

\bibliographystyle{IEEEtranN}
}

\end{document}